\documentclass{article}

\usepackage{iclr2024_conference,times}

\usepackage{tabularx}
\usepackage{microtype}
\usepackage{graphicx}
\usepackage{makecell}
\usepackage{subcaption}
\usepackage{siunitx}
\usepackage{booktabs} 
\usepackage[version=4]{mhchem}
\usepackage{tcolorbox}
\usepackage{color}
\usepackage{colortbl}
\everypar{\looseness=-1}
\usepackage{algorithm}
\usepackage{wrapfig}
\usepackage{algorithmicx}
\usepackage{enumitem}
\usepackage{amssymb}
\usepackage[noend]{algpseudocode}

\newcommand*\Let[2]{\State #1 $\gets$ #2}
\algrenewcommand\alglinenumber[1]{
{\sf\footnotesize\addfontfeatures{Colour=888888,Numbers=Monospaced}#1}}
\algrenewcommand\algorithmicrequire{\textbf{Input:}}

\graphicspath{{./figures/pngs}{./figures/pngs/generated_molecules}{./figures/pdfs}}

\usepackage{hyperref}

\usepackage{iclr2024_conference}

\usepackage{amsmath}
\usepackage{amssymb}
\usepackage{mathtools}
\usepackage{amsthm}
\usepackage{titlesec}
\titlespacing*{\section}{0pt}{0.1\baselineskip}{0.2\baselineskip}
\everypar{\looseness=-1}

\theoremstyle{plain}
\newtheorem{theorem}{Theorem}[section]

\theoremstyle{definition}

\theoremstyle{remark}

\newcommand{\secondbest}[1]{\textcolor{gray}{\textbf{#1}}}
\newcommand{\best}[1]{\textbf{#1}}

\title{Symphony: Symmetry-Equivariant Point-Centered Spherical Harmonics for 3D Molecule Generation}

\author{Ameya Daigavane$^1$, Song Kim$^1$, Mario Geiger$^{2}$\thanks{Work performed when at Massachusetts Institute of Technology.}, \hspace{0.1em} Tess Smidt$^1$ \\
\texttt{\{ameyad,songk\}@mit.edu}, \texttt{geiger.mario@gmail.com}, \texttt{tsmidt@mit.edu} \\
$^1$Massachusetts Institute of Technology \hspace{0.5em}
$^2$NVIDIA
}

\iclrfinalcopy 

\begin{document}

\newcommand{\E}{\mathbb{E}}
\def\gR{\mathcal{R}}
\def\gS{\mathcal{S}}
\def\gN{\mathcal{N}}
\def\gM{\mathcal{M}}
\def\r{\vec{\mathbf{r}}}
\def\R{{\mathbf{R}}}
\def\T{{\mathbf{T}}}
\def\pT{p_\Theta}
\def\fT{f_\Theta}
\def\lmax{l_{\text{max}}}

\DeclarePairedDelimiterX{\infdivx}[2]{(}{)}{%
  #1\;\delimsize\|\;#2%
}
\newcommand{\infdiv}{D_{\text{KL}}\infdivx}
\newcommand{\infdivl}{D_{\text{l2}}\infdivx}
\DeclarePairedDelimiter{\norm}{\lVert}{\rVert}

\maketitle

\begin{abstract}
We present Symphony, an $E(3)$-equivariant autoregressive generative model for 3D molecular geometries that iteratively builds a molecule from molecular fragments.
Existing autoregressive models such as G-SchNet \citep{gschnet} and G-SphereNet \citep{gspherenet} for molecules utilize rotationally invariant features to respect the 3D symmetries of molecules.
In contrast, Symphony uses message-passing with higher-degree $E(3)$-equivariant features. This allows a novel representation of probability distributions via spherical harmonic signals to efficiently model the 3D geometry of molecules. We show that Symphony is able to accurately generate small molecules from the QM9 dataset, outperforming existing autoregressive models and approaching the performance of diffusion models.
\end{abstract}

\section{Introduction}
\label{sec:intro}

In silico generation of atomic systems with diverse geometries and desirable properties is important to many areas including fundamental science, materials design, and drug discovery \citep{perspective}. The direct enumeration and validation of all possible 3D structures is computationally infeasible and does not in itself lead to useful representations of atomic systems for guiding understanding or design. Machine learning methods offer a promising avenue to explore this landscape by learning to generate 3D molecular structures.

Effective generative models of atomic systems must learn to represent and produce highly-correlated geometries that represent chemically valid and energetically favorable configurations. To do this, they must overcome several challenges: 

\begin{itemize}[leftmargin=1em]
\item The validity of an atomic system is ultimately determined by quantum mechanics. Generative models of atomic systems are trained on 3D structures relaxed through computationally-intensive quantum mechanical calculations. These models must learn to adhere to chemical rules, generating stable molecular structures based solely on examples.

\item The stability of atomic systems hinges on the precise placement of individual atoms. The omission or misplacement of a single atom can result in significant property changes and instability.

\item Atomic systems have inherent symmetries. Atoms of the same element are indistinguishable, so there is no consistent way to order atoms within an atomic system. Additionally, atomic systems lack unique coordinate systems (global symmetry) and recurring geometric patterns occur in a variety of locations and orientations (local symmetry).
\end{itemize}

Taking these challenges into consideration, the majority of generative models for atomic systems operate on point geometries and use permutation and Euclidean symmetry-invariant or equivariant methods. Thus far, two approaches have been emerged as effective for directly generating general 3D geometries of molecular systems: autoregressive models \citep{gschnet,cgschnet,gspherenet,simm2020rl,simm2021symmetryaware} and diffusion models \citep{edm}.

In this work, we introduce Symphony, an autoregressive generative model that uses higher-degree equivariant features and spherical harmonic projections to build molecules while respecting the $E(3)$ symmetries of molecular fragments. Similar to other autoregressive models, Symphony builds molecules sequentially by predicting and sampling atom types and locations of new atoms based on conditional probability distributions informed by previously placed atoms. However, Symphony stands out by using spherical harmonic projections to parameterize the distribution of new atom locations. This approach enables predictions to be made using features from a single `focus' atom, which serves as the chosen origin for that step of the generation process. It allows for the simultaneous prediction of the radial and angular distribution of possible atomic positions in a direct manner without needing to use additional atoms.

To test our proposed architecture, we apply Symphony to the QM9 dataset and show that it outperforms previous autoregressive models and is competitive with existing diffusion models on a variety of metrics. We additionally introduce a metric based on the bispectrum for assessing the angular accuracy of matching generated local environments to similar environments in training sets. Finally, we demonstrate that Symphony can generate valid molecules at a high success rate, even when conditioned on unseen molecular fragments.

\begin{figure*}[thbp]
    \centering
    \includegraphics[width=\textwidth]{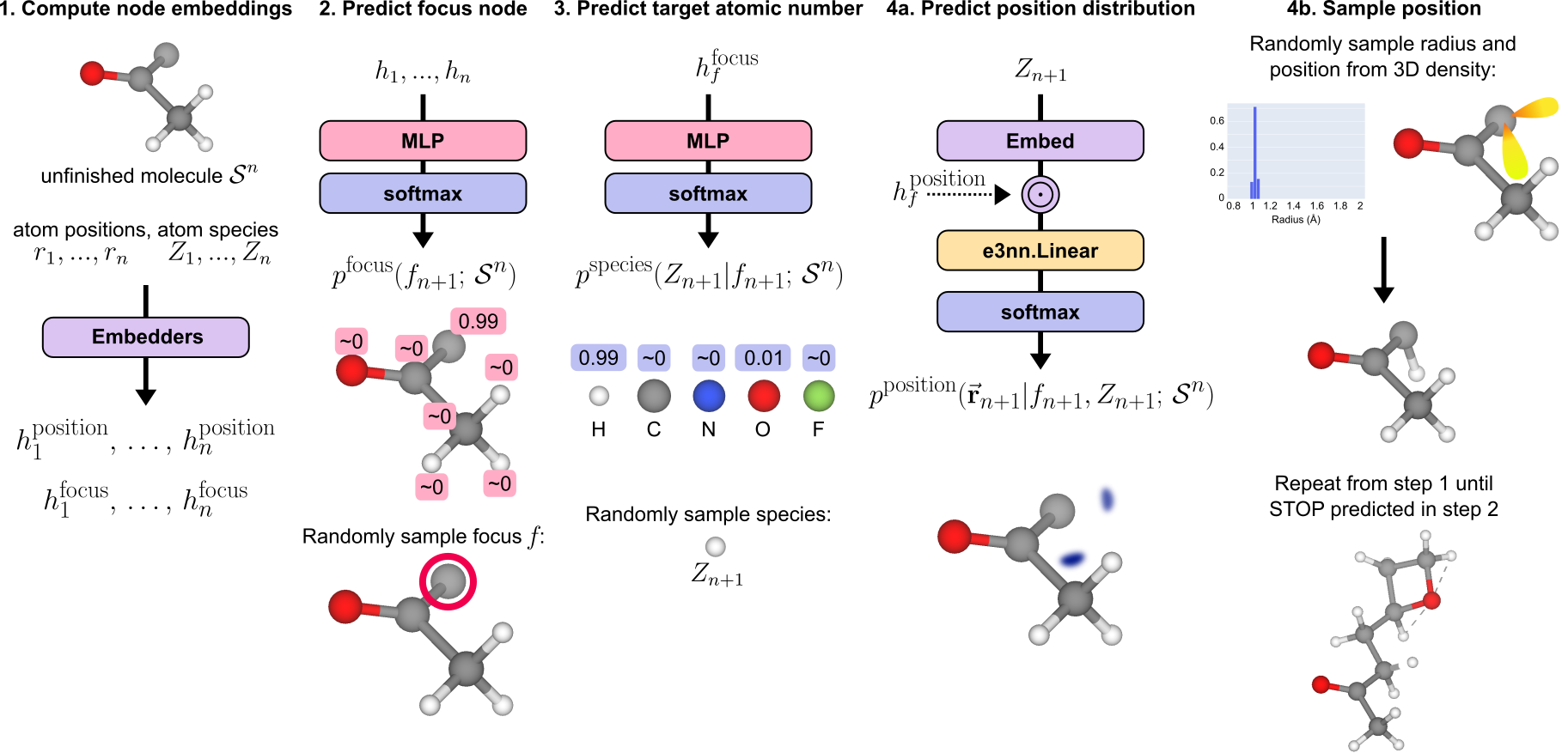}
    \caption{One iteration of the Symphony molecular generation process, in which one atom
    is sampled given the positions and atom types of an unfinished molecular fragment 
    $\gS^n$.
    The complete molecule after all iterations is shown in the bottom right of the figure.}
    \label{fig:model_construction}
\end{figure*}



\section{Background}
\label{sec:background}

\textbf{$E(3)$-Equivariant Features}: We say a $E(3)$-equivariant feature
$z \in \mathbb{R}^{2l + 1}$ transforms as the irreducible representation
$l$ under rotation $\R$ and translation $\T$:
$$\begin{aligned}
    z \xrightarrow{\R, \T} D^l(\R)^T z
\end{aligned}$$
where $D^l$ is the irreducible representation of $SO(3)$ of degree $2l + 1$.
$D^l(\R) \in \mathbb{R}^{(2l + 1) \times (2l + 1)}$ is referred to as the Wigner D-matrix of the rotation $\R$. As $D^0(\R) = 1$ and $D^1(\R) = \R$, invariant `scalar' features correspond to degree $l = 0$ features, while `vector' features correspond to $l = 1$ features. Note that these features are invariant under translation $\T$.

\textbf{Spherical Harmonics}:
The real spherical harmonics $Y_{l,m}(\theta, \phi)$ are a set of real-valued orthogonal functions defined on the sphere $S^2$, indexed by two integers $l$ and $m$ such that $l \geq 0, |m| \leq l$. 
Here $\theta$ and $\phi$ come from the notation for spherical coordinates, where $r$ is the distance from an origin, $\theta \in [0, \pi]$ is the polar angle and $\phi \in [0, 2\pi)$ is the azimuthal angle. The relation between Cartesian and spherical coordinates is given by: $ x = r \sin \theta \cos \phi, y = r \sin \theta \sin \phi, z = r \cos \theta$.

\begin{figure*}[h]
    \centering
    \includegraphics[width=0.8\textwidth]{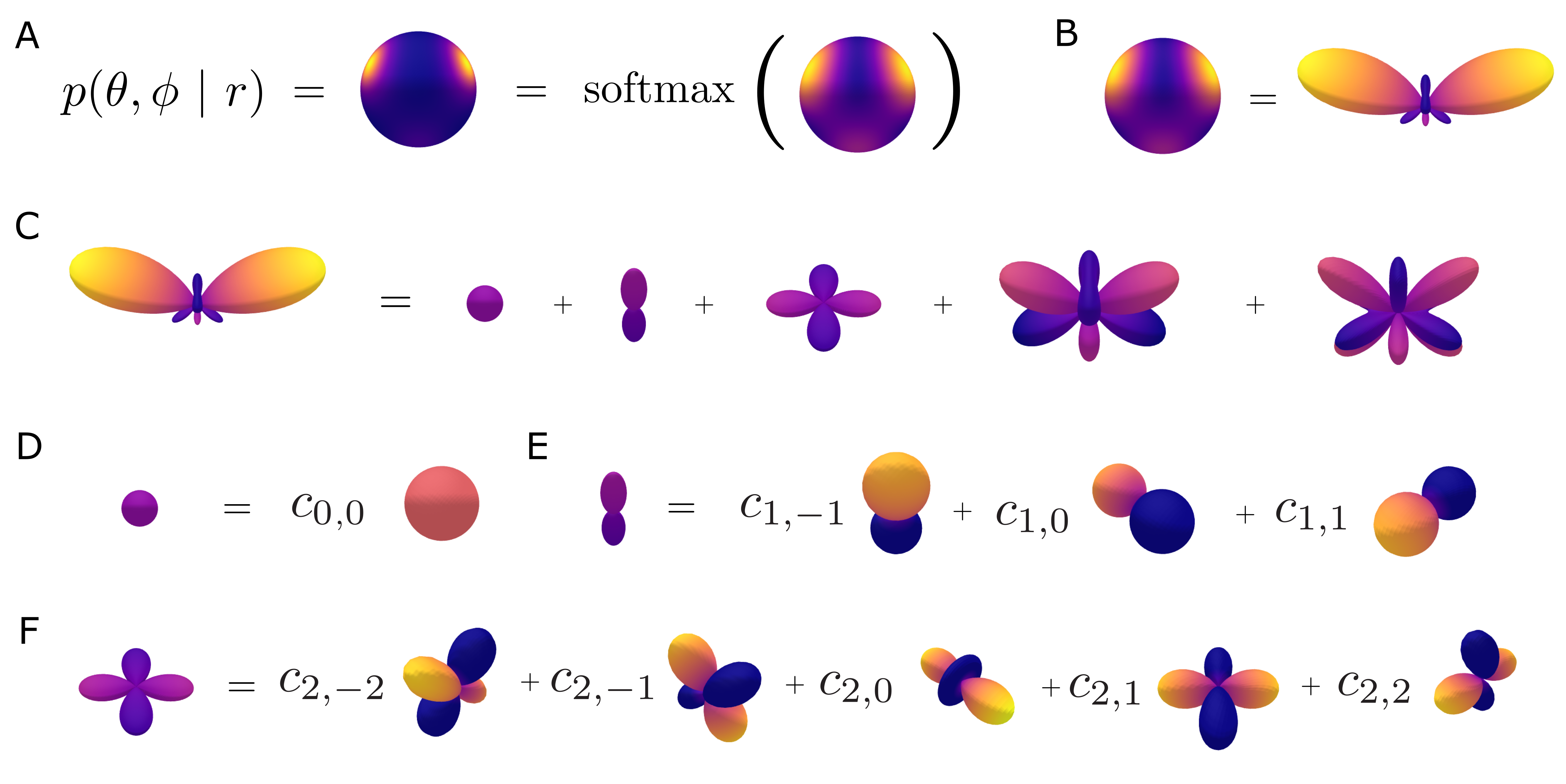}
    \caption{Symphony represents the atom position logits using radial shells of spherical signals. (A)
    illustrates an example angular distribution for a given radial shell prior to applying the softmax function. The softmax enhances signal peaks and prediction precision. (B) presents two ways to plot the signals for a single shell: as a colored sphere, or as a surface where distance from the origin represents signal magnitude, enhancing peak visibility. (C) breaks the pre-activated signal into contributions from $l=0$ to $l=4$ spherical harmonics. (D), (E), and (F) further break these contributions into the $2l+1$ spherical harmonics for $l=0, 1$ and $2$.}
    \label{fig:spherical_harmonic_decomposition}
\end{figure*}

$l$ corresponds to an angular frequency: the higher the $l$, the more rapidly $Y_{l,m}$ changes over $S^2$. This can intuitively be seen by looking at the functional form of the spherical harmonics. In their Cartesian form, the spherical harmonics are proportional to simple polynomials. In one common choice of basis, $l=0$ is proportional to $1$, $l=1$ is proportional to $(x, y, z)$ and $l=2$ is proportional to $(xy, yz, 2z^2 - x^2 -y^2, zx, x^2-y^2)$, as seen in Figure~\ref{fig:spherical_harmonic_decomposition}D-F.



One important property of the spherical harmonics is that they can be used to create $E(3)$-\emph{equivariant} features.
Let $Y_l(\theta, \phi)= [Y_{l,-l}(\theta, \phi), \ldots, Y_{l,l}(\theta, \phi)] \in \mathbb{R}^{2l + 1}$ represent the collection of all spherical harmonics with the same $l$. Then, $Y_l(\theta, \phi)$ transforms as an $E(3)$-equivariant feature of degree $l$ under rotation:
$Y_l(\R (\theta, \phi)) = D^l(\R)^T Y_l(\theta, \phi)
$,
where $\R$ is an arbitrary rotation, and $(\theta, \phi)$ is interpreted as the coordinates of a point on $S^2$.

The second important property of the spherical harmonics that we employ is the fact that they form an \emph{orthonormal basis} for functions on the sphere $S^2$. Thus, for any function $f: S^2 \to \mathbb{R}$, we can express $f$ as a linear combination of the $Y_{l,m}$. Formally, there exists unique coefficients $c_l \in \mathbb{R}^{2l + 1}$ for each $l \in \mathbb{N}$, such that
$
f(\theta, \phi) = \sum_{l = 0}^{\infty} {c_l}^T Y_l(\theta, \phi)
$. We term these coefficients $c_l$ as the spherical harmonic coefficients of $f$
as they are obtained by projecting $f$ onto the spherical harmonics.


\section{Methods}

We first describe Symphony, our autoregressive model for 3D molecular structures, with a comparison to prior work in \autoref{sec:related}.

\subsection{Building Molecules Via Sequences of Fragments}

\begin{figure}[htbp]
  \begin{minipage}{0.38\linewidth}
    \vspace{10pt}
    \includegraphics[width=1.1\linewidth]{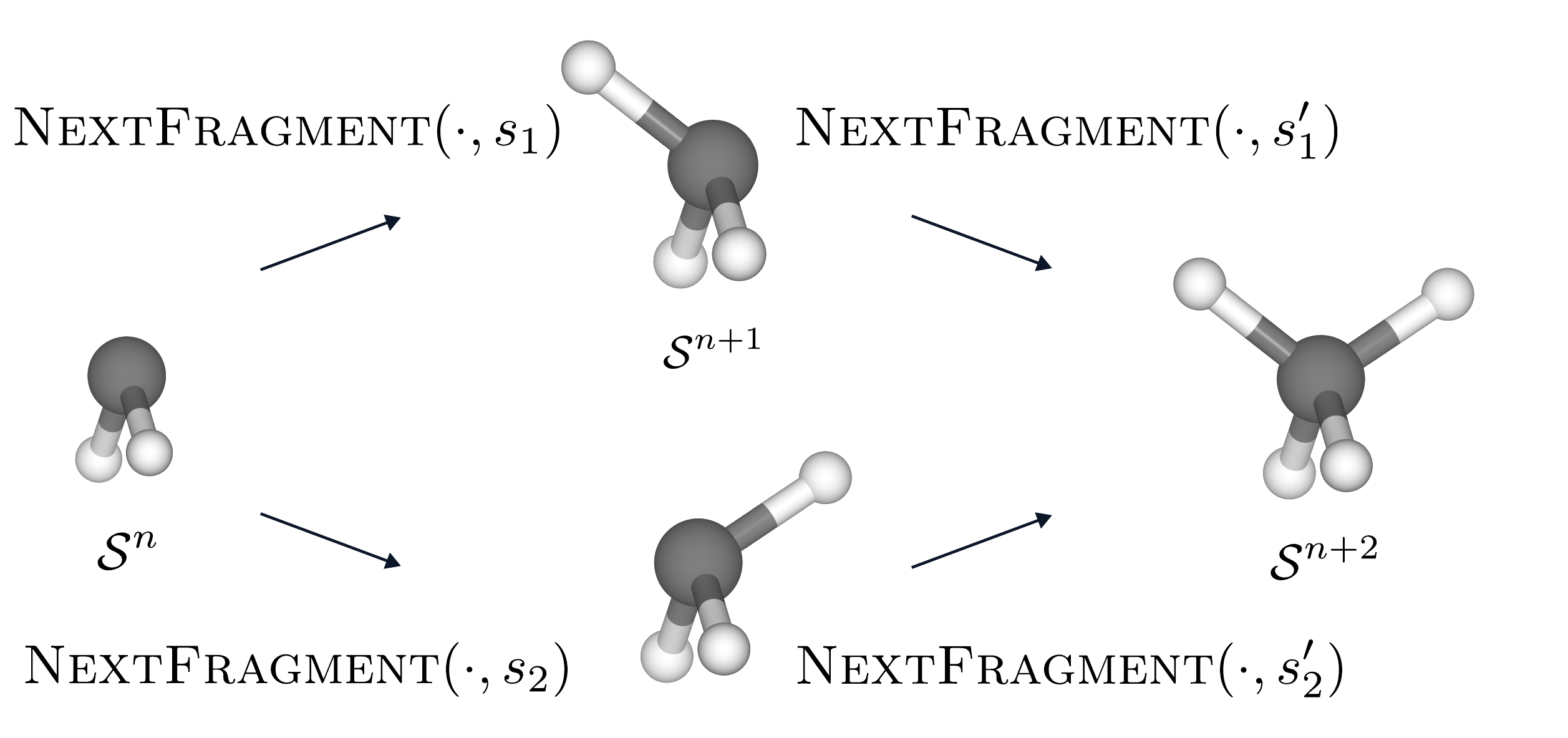}
    \smallskip
    \includegraphics[width=1.0\linewidth]{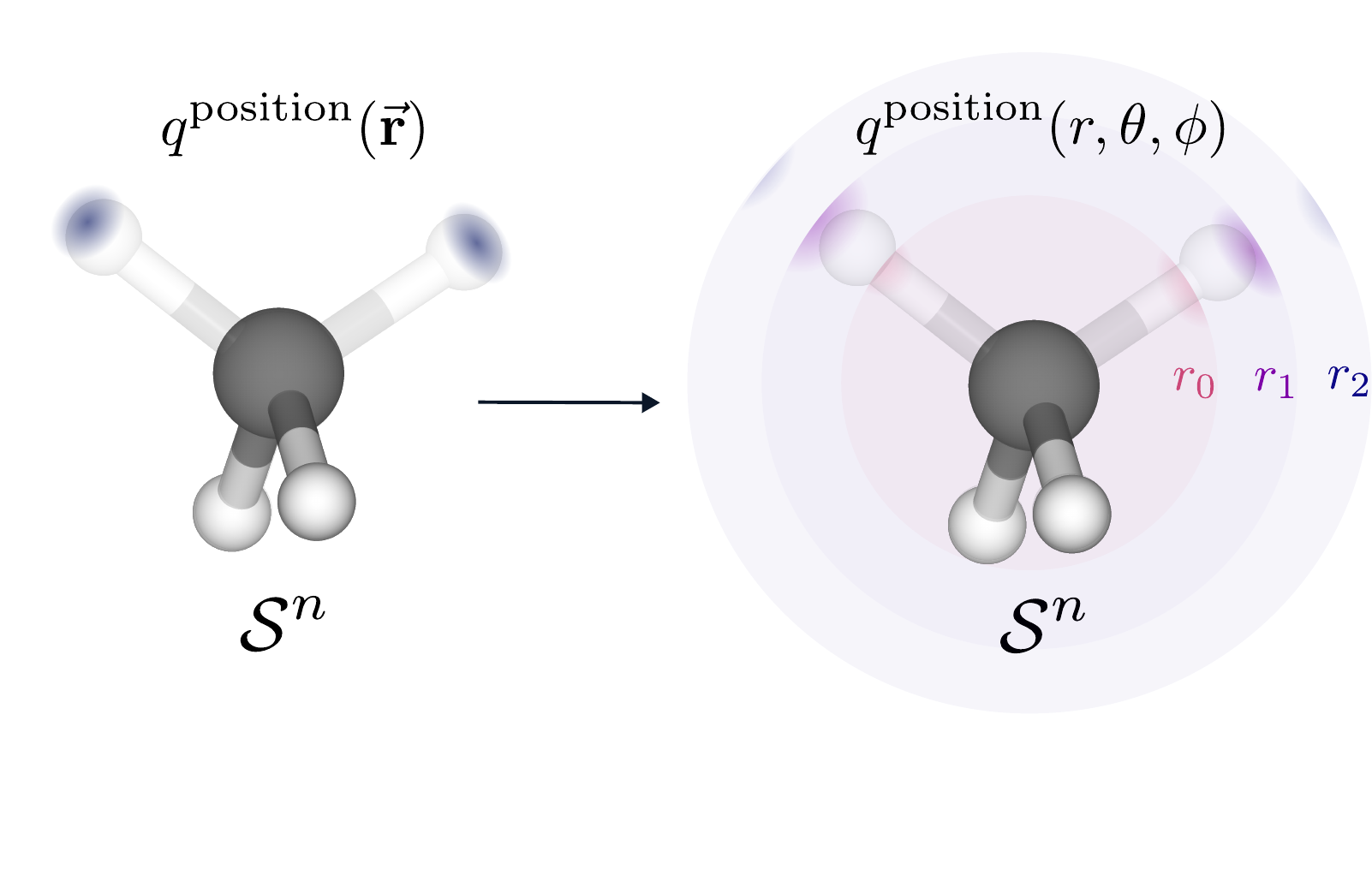}
  \end{minipage}%
  \begin{minipage}{0.62\linewidth}
     \begin{algorithm}[H]
\caption{\textsc{CreateFragmentSequence}}
    \label{alg:createSequence}
    \begin{algorithmic}
    \Require Molecule $\gM$, PRNG Seed $s$
    \State Sample an atom $(\r_1, Z_1)$ from $\gM$ using the PRNG seed $s$.
    \Let{$\gS^{1}$}{$\{(\r_1, Z_1)\}$}
    \Function{\textsc{NextFragment}}{$\gS^{n}, s$}
      \Let{$(f, a)$}{Closest atom pair s.t. $f \in \gS^{n}$ and $a \in \gM - \gS^{n}$}
      \State {(Ties are broken randomly using seed $s$)}
      \Let{$f_{n + 1}$}{The index of the atom $f$ in $\gS^{n}$} 
      \Let{$Z_{n + 1}$}{The atomic number of atom $a$}
      \Let{$\r_{n + 1}$}{The relative position of atom $a$ w.r.t. atom $f$}
      \Let{$\gS^{n + 1}$}{$\gS^{n} \cup \{(\r_{n + 1}, Z_{n + 1})\}$}
      \Let{$s'$}{Update PRNG Seed $s$}
      \State \Return{$(\gS^{n + 1}, s')$}
    \EndFunction
    \For{$n \gets 1 \textrm{ to } |\gM| - 1$}
        \Let{$(\gS^{n + 1}, s)$}{\textsc{NextFragment}($\gS^{n}, s$)} 
    \EndFor
    
    \State \Return $\{\gS^{1}, \ldots \gS^{|\gM|}\}$  
    \end{algorithmic}
    \end{algorithm}
  \end{minipage}
  
  \label{fig:fragment_sequence}
      \caption{(Top) Fragments from $\textsc{CreateFragmentSequence}$
    applied to methane (CH4).
    From $\gS^{n}$, there are thus two valid positions to place the next H atom around the focus $f_{n + 1}$. (Bottom-Left) The true probability distribution $q^\text{position}(\r)$ is smoothly projected onto (Bottom-Right) radial shells of spherical signals according to \autoref{eqn:true_distribution}. All the radial shells show the same angular distribution, but the shell corresponding to $r_1$ is the most probable.}
\end{figure}

First, we create sequences of fragments using molecules from the training set
via~\hyperref[alg:createSequence]{$\textsc{CreateFragmentSequence}$}.
Given a molecule
$\gM$ and random seed $r$, $\textsc{CreateFragmentSequence}$
constructs a sequence of increasingly larger fragments $\{\gS^{1}, \ldots \gS^{|\gM|}\}$
such that $|\gS^{n}| = n$ for all $n \in \{1, \ldots, |\gM|\}$ and $\gS^{|\gM|} = \gM$
exactly.
Of course, there are many ways to
create such sequences of fragments; ~\hyperref[alg:createSequence]{$\textsc{CreateFragmentSequence}$} simply builds a molecule via a minimum spanning tree.

Symphony attempts to recreate this sequence step-by-step, learning the (probabilistic) mapping $\gS^{n}
\to \gS^{n + 1}$. In particular, we ask
Symphony to predict the focus node $f_{n + 1}$, the target atomic number $Z_{n + 1}$ and
the target position $\r_{n + 1}$, providing feedback at every step. 

\subsection{Handling the Symmetries of Fragments}

Here, we highlight several challenges that arise because $\gS^{n}$ must be treated as an unordered set
of atoms that live in 3D space.
In particular, let $\R \gS^{n} + \T = \{(\R\r_1 + \T, Z_1), \ldots, (\R\r_n + \T, Z_n)\}$ be
the description of the same set of atoms in $\gS^{n}$ but in a coordinate frame rotated by $\R^{-1}$ and translated by $\T^{-1}$. 
Similarly, let $\pi$ be any permutation of $\{1, \ldots, n\}$ and $\pi \gS^{n} = \{(\r_{\pi(1)}, Z_{\pi(1)}), \ldots, (\r_{\pi(n)}, Z_{\pi(n)})\}$. 
Fundamentally, $\R \gS^{n} + \T$, $\gS^{n}$ and $\pi
\gS^{n}$ all represent the same set of atoms.
Thus, we would like Symphony to naturally accommodate the 
symmetries of fragment $\gS^n$, under the group $E(3)$ of
Euclidean
transformations consisting of all rotations $\R$ and
translations $\T$, and the group of all permutations of 
constituent atoms. Formally, we wish to have:
\begin{itemize}
    \item \textbf{Property (1)}: The focus distribution
    $p^\text{focus}$ and the target species distribution 
    $p^\text{species}$ should be \emph{$E(3)$-invariant}:  \label{prop:1} 
    \begin{align}
        p^\text{focus}(f_{n + 1}; \R \gS^{n} + \T) &= p^\text{focus}(f_{n + 1}; \gS^{n}) \\       p^\text{species}(Z_{n + 1} \ | \ f_{n + 1}; \R \gS^{n} + \T) &= p^\text{species}(Z_{n + 1} \ | \ f_{n + 1}; \gS^{n})
    \end{align} 
        
    \item \textbf{Property (2)}: The target position distribution $p^\text{position}$ should be \emph{$E(3)$-equivariant}:
    \label{prop:2}
    \begin{align}
        p^\text{position}(\R \r_{n + 1}  + \T  \ | \ f_{n + 1}, Z_{n + 1}; \R \gS^{n} + \T)
        = p^\text{position}(\r_{n + 1} \ | \  f_{n + 1}, Z_{n + 1}; \gS^{n})
    \end{align}

    \item \textbf{Property (3)}: With respect to the ordering of atoms in $\gS^n$, the map $p^\text{focus}$ should be permutation-equivariant while $p^\text{species}$ and
    $p^\text{position}$ should be permutation-invariant. \label{prop:3}
\end{itemize}

We represent $p^\text{focus}, p^\text{species}$ and $p^\text{position}$ as
probability distributions because
there may be multiple valid choices of focus $f_{n + 1}$, species $Z_{n + 1}$ and
position $\r_{n + 1}$.

\subsection{The Design of Symphony}
\label{sec:symphony-design}
The overall working of Symphony is shown graphically in \autoref{fig:model_construction}.
Symphony first computes atom embeddings via an $\textsc{Embedder}$.
Here, we assume that
$\textsc{Embedder}(\gS^n) = \{h_{v,l} \ | \ v \in \gS^n, 0 \leq l \leq \lmax \}$ returns a
set of $E(3)$-equivariant features $h_v,l$ of degree $l$ upto a predefined degree $\lmax$, for each atom 
$v$ in $\gS^n$. 
In \autoref{thm:equivariance-proof}, we show that Symphony can guarantee
\hyperref[prop:1]{\textbf{Properties (1), (2)}} and \hyperref[prop:3]{\textbf{(3)}} as long as $\textsc{Embedder}$ is permutation-equivariant and $E(3)$-equivariant. 

From \hyperref[prop:1]{\textbf{Property (1)}}, $p^\text{focus}$ and $p^\text{species}$ should be invariant 
under rotation and translations of $\gS^n$. Since the atom types and the atom indices are discrete sets, we can represent both of these distributions as a vector of probabilities.
Thus, we compute $p^\text{focus}$ and $p^\text{species}$
by applying a multi-layer perceptron $\textsc{MLP}$ on only the rotation and translation
invariant features of $\textsc{Embedder}(\gS^n)$:
\begin{align}
    p^\text{focus}(f_{n + 1} ; \gS^n)  &= \textsc{MLP}(\textsc{Embedder}(\gS^n)_{f_{n + 1},0})
    \\
    p^\text{species}(Z_{n + 1} \ | \ f_{n + 1} ; \gS^n) &= \textsc{MLP}(\textsc{EmbedAtomType}(Z_{n + 1}) \cdot \textsc{Embedder}(\gS^n)_{f_{n + 1},0})
    \label{eqn:focus_and_atom_type}
\end{align}
Alongside the node-wise probabilities for $p^\text{focus}$, we also predict a global  STOP probability, indicating that no atom should be added.

On the other hand, \hyperref[prop:2]{\textbf{Property (2)}} shows that $p^\text{position}$ transforms
non-identically under rotations and translations.
We describe a novel parametrization of 3D probability densities such as $p^\text{position}$
with spherical harmonic projections.

The position $\r$ is represented by spherical coordinates $(r, \theta, \phi)$ where $r$ is
the distance from the focus $f$, $\theta$ is the polar angle and $\phi$ is the azimuthal 
angle.
Any probability distribution $p^\text{position}$ over positions must satisfy the 
normalization and non-negativity constraints:
$
\int_{\Omega} p^\text{position}(r, \theta, \phi) \ dV = 1 \ \text{and} \
p^\text{position}(r, \theta, \phi) \geq 0
$
where $dV = r dr \sin \theta d\theta d\phi$ is the volume element and $\Omega = \{ r \in [0, \infty),
\theta \in [0, \pi], \phi \in [0, 2\pi)\}$ represents all space in spherical coordinates.
Since these constraints are hard to incorporate directly into a neural network, we predict 
the unnormalized logits $f^\text{position}(r, \theta, \phi)$ instead, and take the softmax 
over all space:
$
p^\text{position}(r, \theta, \phi) = \frac{1}{Z} \exp{f^\text{position}(r, \theta, \phi)}
$
To model these logits, we
first discretize the radial component $r$ into a set of discrete values. We choose $64$ uniformly spaced values from $0.9$A to $2.0$A, which covers all of the bond lengths in QM9.
For each fixed value of $r$, we obtain a function on the sphere $S^2$, which we represent
in the basis of spherical harmonic functions $Y_{l,m}(\theta, \phi)$, as described in 
\autoref{sec:background} and similar to the construction of \citet{cohen2015harmonic}. As we have a radial component $r$ here, the coefficients 
$c_l$ also depend on $r$:
$$
f^\text{position}(r, \theta, \phi \ | \ f_{n + 1}, Z_{n + 1};  \gS^{n}) = \sum_{l = 0}^{\infty}
c_l(r;  f_{n + 1}, Z_{n + 1}, \gS^{n})^T Y_l(\theta, \phi)
$$
Symphony predicts these coefficients $c_l$ from the degree $l$ features of the focus 
node $\textsc{Embedder}(\gS^n)_{f_{n + 1},l}$, and the embedding of the target species
$Z_{n + 1}$:
\begin{align}
    \label{eqn:position_coeffs}
    c_l(r;  f_{n + 1}, Z_{n + 1}, \gS^{n})
    = \textsc{Linear}(\textsc{Embedder}(\gS^n)_{f_{n + 1},l} \otimes \textsc{EmbedAtomType}(Z_{n + 1}))
\end{align}
By explicitly modelling the probability distributions $p^\text{focus}, p^\text{species}$ 
and $p^\text{position}$, Symphony learns to represent all possible options of completing
$\gS^{n}$ into a valid molecule.

\subsection{Bypassing the Angular Frequency Bottleneck}
For computational reasons, we are often limited to using a finite number of spherical 
harmonic projections (ie, up to some $\lmax$). Due to the way the spherical harmonics are 
constructed, this means we can only represent signals upto some angular frequency. For 
example, to represent a signal on the sphere with peaks separated by $d$ radians, we need 
spherical harmonic projections with $\lmax \geq \frac{2\pi}{d}$. This is similar to issues 
faced when using the first few terms of the Fourier series; we cannot represent high 
frequency components.
To bypass the bottleneck of angular frequency, we propose using \emph{multiple} channels of 
spherical harmonic projections, which are then summed over after a non-linearity:
$
f^\text{position}(r, \theta, \phi;  \gS^{n}) = \log \sum_{\text{channel ch}} \exp{
\sum_{l = 0}^{\infty} c^{\text{ch}}_{l}(r; \gS^{n})^T Y_l(\theta, \phi)}
$. See \autoref{sec:channels-example} for a concrete example
where adding multiple channels effectively increases the
angular frequency capacity of our model.
For Symphony, we find that $2$ channels is sufficient, as demonstrated in

\subsection{Training and Inference}
We utilize teacher forcing to train Symphony.
At training time, the true focus $f_{n + 1}$ and atomic number $Z_{n + 1}$
are provided as computed in $\textsc{NextFragment}$.
Thus, no sampling occurs at training time. The true probability distributions $q^\text{focus}$ and $q^\text{species}$ are computed empirically from the set of unfinished atoms and their corresponding neighbors in $\gM$.
The true probability distribution $q^\text{position}$ is computed by smoothly approximating a Dirac delta distribution upto some cutoff frequency $\lmax$ at the target position $\r_{n + 1}$ around the focus atom. Further details about the training process and representing  Dirac delta distributions are provided in \autoref{sec:training-details} and \autoref{sec:delta-functions}.
\begin{align}
    q^\text{position}(\r) = \frac{1}{Z} \exp{\left(-\frac{\norm{\r} - \norm{\r_{n + 1}}}{2\sigma_{\text{true}}^2} \cdot \delta_{\lmax}\left(\hat{\textbf{r}} - \hat{\textbf{r}}_{n + 1}\right)\right)}
    \label{eqn:true_distribution}
\end{align}
At inference time, both the focus $f_{n + 1}$ and atomic number $Z_{n + 1}$
are sampled from $p^\text{focus}(\cdot; \gS^n)$ and 
$p^\text{species}(\cdot | f_{n + 1}; \gS^n)$ respectively. These are used to sample $\r_{n + 1}$ from $p^\text{position}(\cdot | f_{n + 1}, Z_{n + 1}; \gS^{n})$. Molecules are generated by starting from an initial
fragment $\gS^{1}$, and repeatedly sampling from $p^\text{focus}$,  
$p^\text{species}$ and $p^\text{position}$
until a STOP is predicted or $N_{\text{max}} = 35$ iterations have 
occurred, based on the maximum size of molecules
in the QM9 dataset as $30$ atoms.

\subsection{Relation to Prior Work}
\label{sec:related}
Most methods for 3D molecular structure generation fall into one of two broad categories: autoregressive and end-to-end models.
G-SchNet \citep{gschnet,cgschnet} and G-SphereNet \citep{gspherenet} were the first successful attempts at autoregressive generation of molecular structures. 

G-SchNet uses the SchNet framework \citep{schnet} to perform message-passing with rotationally invariant
features and compute node embeddings. A focus node is then selected as the center of a 3D grid. All of the atoms in the current fragment then vote on where to place the next atom within this grid by specifying a radial distance to the next atom.
Because of the use of only rotationally invariant features, at least three atoms are needed to be present in the current fragment to specify the exact position of the next atom without any degeneracy due to symmetry; this procedure is called \emph{triangulation}. This requires several additional tokens to break symmetry.
Similarly, G-SphereNet learns a normalizing flow to perform a triangulation procedure once there are atleast $3$ atoms in $\gS^{n}$.

We wish to highlight two observations that guided the development of Symphony:
\begin{itemize}[leftmargin=1em]
    \item Rotationally invariant features centered at a single point  cannot capture the orientations of geometrical
    motifs \citep{incompleteness}. To handle the degeneracies inherent when using 
    rotationally invariant features to predict positions, G-SchNet uses unphysical 
    auxiliary tokens (which are multiple spatial positions that are not atoms) to break symmetry. 
    \item G-SchNet queries all of the atoms in $\gS^{n}$ at each iteration, which means 
    distant atoms can have an undue influence when placing the next atom. Similarly, G-
    SphereNet predictions are not a smooth function of the input fragment; when the input 
    is perturbed slightly, the choice of atoms used in the triangulation procedure can 
    change drastically. 
\end{itemize}

Recently, $E(3)$-equivariant neural networks that build higher-degree $E(3)$-equivariant
features have demonstrated improved performance on a wide range of atomistic tasks 
\citep{nequip,e3nn,tm23}.
Our key contribution is to show the benefit of higher-degree 
$E(3)$-equivariant features for the \emph{molecular generation} task
allowing for a novel
parametrization of 3D probability distributions using spherical harmonic
projections.
\citet{simm2021symmetryaware} also uses spherical harmonic projections with a 
single channel for molecule generation, but trained with reinforcement learning, and sampled using rejection sampling.
We discuss these details in \autoref{sec:position-distributions}.

Among end-to-end generation methods, \citet{edm} developed EDM, a state-of-the-art $E(3)$-equivariant diffusion model.
EDM significantly outperformed the previously proposed $E(3)$-equivariant normalizing flow (ENF) models for molecule generation
\citep{enf}.
EDM learns to gradually denoise a initial configuration of atoms into a
valid molecular structure.
Both EDM and ENF are built on the $E(n)$-Equivariant Graph Neural Networks
\citep{egnn} framework which can utilize only
scalar and vector features (and interactions between them). MiDi
\citep{vignac2023midi} improves EDM by utilizing bond order information (and hence, a 2D
molecular graph to compare to), which we do not assume access to here.
While expressive, diffusion models are expensive to train, requiring $\approx 3.5\times$ more
training on the QM9 dataset to outperform autoregressive models.
Unlike autoregressive models,
diffusion models do not flexibly allow for completion of molecular
fragments, because they are usually trained in setups where all atoms are free to move.
Current diffusion models use
fully-connected graphs where all atoms interact with each other.
This could potentially affect their scalability when building larger molecules. 
On the other hand, Symphony and other autoregressive models use distance cutoffs to restrict interactions and
improve efficiency. In \autoref{sec:edm-radial-cutoff}, we find that EDM
with distance cutoffs performs quite poorly.

Furthermore, diffusion models are significantly slower to sample from,
because the underlying neural network is invoked $\approx 1000$ times when sampling a
single molecule. Flow matching \citep{lipman2023flow} has also emerged as a competitor for diffusion models for 3D molecule generation \citep{song2023equivariant}, but suffers from the same drawbacks listed above.

\section{Experimental Results}
\label{sec:experiments}

A major challenge with generative modelling is evaluating the quality of generated 3D structures.
Ideally, a generative model should generate physically plausible structures, accurately capture training set statistics and
generalize well to molecules outside of its training set.
We propose a comprehensive set of tests to evaluate Symphony and other generative models along these three aspects.

\subsection{Validity of Generated Structures}
All of the generative models considered here output a set of atoms with 3D coordinates; bonding information is not generated by the model. Before we can use cheminformatics tools designed for molecules, we need to assign bonds between atoms. Multiple algorithms exist for bond order assignment: \texttt{xyz2mol} \citep{xyz2mol}, OpenBabel \citep{openbabel} and a simple lookup table based on empirical pairwise distances in organic compounds \citep{edm}. Here, we perform the first comparison between these algorithms for evaluating machine-learning generated 3D structures. In \autoref{tab:validity_metrics}, we use each of these algorithms to infer the bonds and create a molecule from generated 3D molecular structure.  We declare a molecule as valid if the algorithm could successfully assign bond order with no net resulting charge.
We also measure the uniqueness to see how many repetitions were present in the set of SMILES \citep{smiles} strings of valid generated molecules. Ideally, we want both the validity and the uniqueness to be high. 

While EDM \citep{edm} is still superior on the validity and uniqueness metrics, we find that Symphony performs much better on both validity and uniqueness than existing autoregressive models, G-SchNet \citep{gschnet} and G-SphereNet \citep{gspherenet}, for the \texttt{xyz2mol} and OpenBabel algorithms. Note that the lookup table does not account for aromatic bonds and is quite sensitive to exact bond lengths; we believe this penalizes Symphony due to its coarser discretization compared to EDM and G-SchNet.
Of note is that only \texttt{xyz2mol} finds almost all of the ground truth QM9 structures to be valid.

\begin{table}[h]
\centering
    \begin{tabular}{cccccc}
    \toprule
    Metric $\uparrow$ & QM9 & Symphony & EDM & G-SchNet & G-SphereNet  \\  
    \midrule
    Validity via \texttt{xyz2mol} &  99.99 & \secondbest{83.50} & \best{86.74} &   74.97 &      26.92 \\
    Validity via OpenBabel & 94.60 & \secondbest{74.69} & \best{77.75} & 61.83 & 9.86 \\
    Validity via Lookup Table  & 97.60 & 68.11  & \best{90.77} & \secondbest{80.13} & 16.36 \\
    \midrule
    Uniqueness via \texttt{xyz2mol} &  99.84 &  \secondbest{97.98} &  \best{99.16} & 96.73 &      21.69 \\
    Uniqueness via OpenBabel & 99.97 & \secondbest{99.61} & \best{99.95} & 98.71 & 7.51 \\
    Uniqueness via Lookup Table  & 99.89 & \secondbest{97.68}  & \best{98.64} & 93.20 & 23.29 \\
    \end{tabular}
    \caption{Validity and uniqueness (among valid) percentages of molecules with different bond assignment methods, with \best{best} and \secondbest{second-best} models highlighted.}
    \label{tab:validity_metrics}
\end{table}

Recently, \citet{posebusters} showed that the predicted 3D structures from machine-learned protein-ligand docking models tend to be highly unphysical.
For \autoref{tab:posebusters_metrics}, we utilize their PoseBusters framework to perform the following sanity checks to count how many of the predicted 3D structures are reasonable.
We see that the valid molecules from all models tend to be quite reasonable, with Symphony performing better than all baselines on generating structures with reasonable UFF \citep{uff} energies and respecting the geometry constraints of double bonds. Further details about the PoseBusters tests are provided in \autoref{sec:posebusters-details}.

\begin{table}[h]
\centering
    \begin{tabular}{ccccc}
    \toprule
    Test $\uparrow$ & Symphony & EDM & G-SchNet & G-SphereNet  \\
    \midrule
    All Atoms Connected        &           \secondbest{99.92} &            99.88 &   99.87 &  \best{100.00} \\
    Reasonable Bond Angles     &           99.56 &   \best{99.98} &   \secondbest{99.88} &            97.59 \\
    Reasonable Bond Lengths    &           98.72 &  \best{100.00} &   \secondbest{99.93} &            72.99 \\
    Aromatic Ring Flatness     &          \best{100.00} &           \best{100.00} &   99.95 &            99.85 \\
    Double Bond Flatness       &  \best{99.07} &            \secondbest{98.58} &   97.96 &            95.99 \\
    Reasonable Internal Energy &  \best{95.65} &            94.88 &   \secondbest{95.04} &            36.07 \\
    No Internal Steric Clash   &           98.16 &   \best{99.79} &   \secondbest{99.57} &            98.07 \\
    \end{tabular}
    \caption{Percentage of valid (as obtained from \texttt{xyz2mol}) molecules passing each PoseBusters test.}
    \label{tab:posebusters_metrics}
\end{table}
\vspace{-1em}

\subsection{Capturing Training Set Statistics}
Next, we evaluate models on how well they capture bonding patterns and the geometry of local environments found in the training set molecules.
In previous work~\citep{gspherenet,edm}, models were compared based on how well they capture the true bond length distributions observed in QM9. However, such statistics only deal with pairwise bond lengths and cannot capture the geometry of how atoms are placed relative to each other. Here, we utilize the \emph{bispectrum} \citep{bispectrum} as a rotationally invariant descriptor of the geometry of local environments. 
Given a local environment with a central atom $u$, we first project all of the neighbors of $u$ according to the inferred bonds onto the unit sphere $S^2$. Then, we compute the signal $f$ as a sum of Dirac delta distributions along the direction of each neighbor:
$
f(\hat{\textbf{r}}) = \sum_{v \in N(u)}\delta_{\lmax}\left(\hat{\textbf{r}} - \hat{\textbf{r}}_{vu} \right)
$.
The bispectrum $\mathcal{B}(f)$ of $f$
is then defined as:
$
    \mathcal{B}(f) = \textsc{ExtractScalars}(f \otimes f \otimes f)
$.
Thus, $f$ captures the distribution of atoms around $u$, and the bispectrum $\mathcal{B}(f)$ captures the geometry of this distribution. The advantage of the bispectrum is that it varies smoothly when $f$ is varied and is guaranteed to be rotationally invariant. We compute the bispectrum of local environments with atleast $2$ neighboring atoms. Note that we exclude the pseudoscalars in the bispectra.

For comparing discrete distributions, we use the symmetric Jensen-Shannon divergence (JSD) as employed in \citet{edm}. Given the true distribution $Q$ and the predicted distribution $P$, the Jensen-Shannon divergence between them is defined as: $D_{JS}(Q \, \| \, P) = \frac12 D_{KL}\left(Q \, \| \, M\right) + \frac12 D_{KL}\left(P \, \| \, M \right)$ where $D_{KL}$ is the Kullback–Leibler divergence and $M = \frac{Q+P}{2}$ is the mean distribution. For continuous distributions, estimating the Jensen-Shannon divergence from samples is tricky without further assumptions on the distributions. Instead, we use the Maximum Mean Discrepancy (MMD) score from \citet{gspherenet} instead to compare samples from continuous distributions. The MMD score is the distance between means of features computed from samples from the true distribution $Q$ and the predicted distribution $P$.  A model with a smaller MMD score captures the true distribution of samples better. We provide details about the MMD score in \autoref{sec:mmd-details}.

From \autoref{tab:training_set_statistics_matching} we see that Symphony and other autoregressive models struggle to match the bond length distribution of QM9 as well as EDM. This is the case except for the single C-H and single N-H bonds. On the bispectra, however, Symphony attains the lowest MMD for several environments. To gain some intuition for these MMD numbers, we also plotted the bond length distributions, samples of the bispectra, atom type distributions and other statistics in \autoref{sec:additional-analyses} for each model.

\begin{table}[h]
    \begin{subtable}[h]{\textwidth}
        \centering
        \begin{tabular}{ccccc}
        \toprule
        MMD of Bond Lengths $\downarrow$ & Symphony & EDM & G-SchNet & G-SphereNet  \\
        \midrule
        C-H: 1.0 &           0.0739 &  \textbf{0.0653} &  0.3817 &           0.1334 \\
        C-C: 1.0 &           0.3254 &  \textbf{0.0956} &  0.2530 &           1.0503 \\
        C-O: 1.0 &           0.2571 &  \textbf{0.0757} &  0.5315 &           0.6082 \\
        C-N: 1.0 &           0.3086 &  \textbf{0.1755} &  0.2999 &           0.4279 \\
        N-H: 1.0 &  \textbf{0.1032} &           0.1137 &  0.5968 &           0.1660 \\
        C-O: 2.0 &           0.3033 &  \textbf{0.0668} &  0.2628 &           2.0812 \\
        C-N: 1.5 &           0.3707 &  \textbf{0.1736} &  0.5828 &           0.4949 \\
        O-H: 1.0 &           0.2872 &           0.1545 &  0.7899 &  \textbf{0.1307} \\
        C-C: 1.5 &           0.4142 &  \textbf{0.1749} &  0.2051 &           0.8574 \\
        C-N: 2.0 &           0.5938 &  \textbf{0.3237} &  0.4194 &           2.1197 \\
        \end{tabular}
    \end{subtable}
    \begin{subtable}[h]{\textwidth}
        \centering
        \begin{tabular}{ccccc}
        \toprule
        MMD of Bispectra  $\downarrow$ & Symphony & EDM & G-SchNet & G-SphereNet  \\
        \midrule
        C: C2,H2    &           0.2165 &  \textbf{0.1003} &           0.4333 &           0.6210 \\
        C: C1,H3    &           0.2668 &  \textbf{0.0025} &           0.0640 &           1.2004 \\
        C: C3,H1    &  \textbf{0.1111} &           0.2254 &           0.2045 &           1.1209 \\
        C: C2,H1,O1 &  \textbf{0.1500} &           0.2059 &           0.1732 &           0.8361 \\
        C: C1,H2,O1 &           0.3300 &           0.1082 &  \textbf{0.0954} &           1.6772 \\
        O: C1,H1    &           0.0282 &           0.0056 &           0.0487 &  \textbf{0.0030} \\
        C: C2,H1,N1 &  \textbf{0.1481} &           0.1521 &           0.1967 &           1.3461 \\
        C: C2,H1    &           0.2525 &  \textbf{0.0468} &           0.1788 &           0.2403 \\
        C: C1,H2,N1 &           0.3631 &           0.2728 &  \textbf{0.1610} &           0.9171 \\
        N: C2,H1    &  \textbf{0.0953} &           0.2339 &           0.2105 &           0.6141 \\
        \end{tabular}
    \end{subtable}
    \begin{subtable}[h]{\textwidth}
        \centering
        \begin{tabular}{ccccc}
        \toprule
        Jensen-Shannon Divergence $\downarrow$ & Symphony & EDM & G-SchNet & G-SphereNet  \\
        \midrule
        Atom Type Counts &   0.0003 &  \textbf{0.0002} &  0.0011 &     0.0026 \\
        Local Environment Counts &  \textbf{0.0039} &  0.0057 &  0.0150 &     0.1016 \\
        \end{tabular}
    \end{subtable}
    \caption{Comparing statistics of generated molecules to those found in QM9. (Top): The MMD of bond lengths for the $10$ most frequent bonds. The notation `X-Y: T' means that a X atom was bonded to a Y atom with a bond of type T. (Middle): The MMD of bispectra for the $10$ most occurring local environments. The notation `X: Y$n$,Z$m$' means that an X atom was the central atom, surrounded by $n$ Y atoms and $m$ Z atoms. (Bottom): The JSD of occurrence counts for atom types and local environments. $\downarrow$ indicates that lower is better for the metrics.}
    \label{tab:training_set_statistics_matching}
\end{table}
\vspace{-0.5em}

\subsection{Generalization Capabilities}
All of the metrics discussed so far can be maximized by simply 
memorizing the training set molecules. Now, we propose a new metric to 
evaluate how well the models have actually learned to generate valid 
chemical structures. We compare models by asking them to complete 
fragments of $1000$ unseen molecules from the
test set, with one hydrogen atom removed.
We then check how many
final molecules were deemed valid.
Since the valid completion rate (VCR)
depends heavily on the quality of the model,
we compute the valid 
completion rate for fragments of molecules
from the training set as
well. If the performance is significantly
different between the two sets
of fragments,
this indicates that the models do not
generalize well.
Diffusion models such as EDM are more
challenging to evaluate for this task, since
we would need a way to fix the initial set of
atoms, so we compare only Symphony and
G-SchNet. Encouragingly, both models are able
to generalize well to unseen fragments, but
Symphony's overall completion rate is higher
for both seen and unseen fragments.
We notice that the performance of Symphony on this task seems to decrease as training progresses, the reason for which remains unclear.

\begin{table}[h]
\centering
    \begin{tabular}{ccccc}
    \toprule
     Valid Completion Rate $\uparrow$ & \makecell{ Symphony \\ 500K steps } & \makecell{ Symphony \\ 800K steps } & \makecell{ Symphony \\ 1000K steps } & G-SchNet  \\
    \midrule
    Training: $\text{VCR}_\text{train}$ & \textbf{98.53} & 96.65 & 95.57 & 97.91 \\
    Testing: $\text{VCR}_\text{test}$ & \textbf{98.66} & 96.30 & 95.43 & 98.15  \\
    \end{tabular}
    \caption{Comparing the difference between fragment completion rates on (seen) training and (unseen) testing fragments with one hydrogen removed.}
    \label{tab:generalization_metrics}
\end{table}
\vspace{-0.5em}

\subsection{Molecule Generation Throughput}
One of the major advantages of autoregressive models (such as Symphony) over diffusion models (such as EDM) is significantly faster inference speeds. As measured on a single NVIDIA RTX A5000 GPU, Symphony's inference speed is 0.293 seconds/molecule, compared to EDM's 0.930 sec/mol.
Symphony is much slower than existing autoregressive models (G-SchNet is at 0.011 sec/mol, and G-SphereNet 0.006) because of the additional tensor products for generating higher-degree $E(3)$-equivariant features, but is still approximately $3\times$ faster than EDM.  However, our sampler is currently bottlenecked by some of the limitations of JAX \citep{jax2018github}; we believe that Symphony's inference speed reported here can be significantly improved to match its training speed.



\section{Conclusion}
We have proposed Symphony, a new method to autoregressively generate 3D molecular geometries
with spherical harmonic projections and higher-degree $E(3)$-equivariant features.
We show promising results on molecular generation and completion, relative to existing autoregressive models.
However, one drawback of our current formulation is that the discretization of our radial components is too coarse, so our bond length distributions are not as accurate as EDM or G-SchNet. This affects our validity when using lookup tables to assign bond orders as they are particularly sensitive to exact bond lengths.
Further, Symphony incurs increased computational cost due to the use of tensor products to create higher degree $E(3)$-equivariant features. As a highlight, Symphony is trained on only $\approx 80$ epochs, while G-SchNet and EDM are trained for $330$ and $1100$ epochs respectively. Further exploring the data efficiency of Symphony remains to be seen. In the future, we plan to explore normalizing flows to smoothly model the radial distribution without any discretization, and placing entire local environment motifs at once which would speed up generation.

\clearpage

\section{Reproducibility Statement}

Our JAX code containing
all of the data preprocessing,
model training and evaluation
metrics is available at \url{https://github.com/atomicarchitects/symphony}.

\autoref{sec:training-details} describes the hyperparameters used in the training process for Symphony.
Details about the metrics used can be found in \autoref{sec:metrics-details}.
\autoref{sec:data-details} contains all of the information regarding the QM9 dataset used in this work. Further information about the baseline models and the sampled structures can be found in \autoref{sec:baseline-details}. 
\section{Ethics Statement}
Generative models for molecules such as Symphony have the potential to be used for
discovering novel drugs and useful catalysts.
While harmful uses of such generative models exist,
the synthesis of a molecule given only its 3D geometry is still extremely challenging.
Thus, we do not anticipate any negative consequences of our research.

\subsubsection*{Acknowledgments}
Ameya Daigavane was supported by the National Science Foundation under Cooperative Agreement PHY-2019786 (The NSF AI
Institute for Artificial Intelligence and Fundamental Interactions),
and the NSF Graduate Research Fellowship program. Song Kim was supported by Analog Devices as an Undergraduate Research and Innovation Scholar
in the MIT Advanced Undergraduate Research Opportunities Program (SuperUROP). Mario Geiger and Tess Smidt were supported by the Integrated Computational and Data Infrastructure (ICDI) program of the U.S. Department of Energy, grant number DE-SC0022215. The authors acknowledge the MIT SuperCloud and Lincoln Laboratory Supercomputing Center for providing HPC resources that have contributed to the research results reported within this paper.

\bibliography{references}
\bibliographystyle{iclr2024_conference}

\newpage
\appendix
\onecolumn
\section*{Appendix}

\section{Additional Analyses}
\label{sec:additional-analyses}
For all of the analyses performed in this section, we used all the valid molecules for each model as computed by \texttt{xyz2mol}.

\subsection{Bispectra of Local Environments in Sampled Molecules}
\begin{figure}[h]
    \centering
    \includegraphics[width=\textwidth]{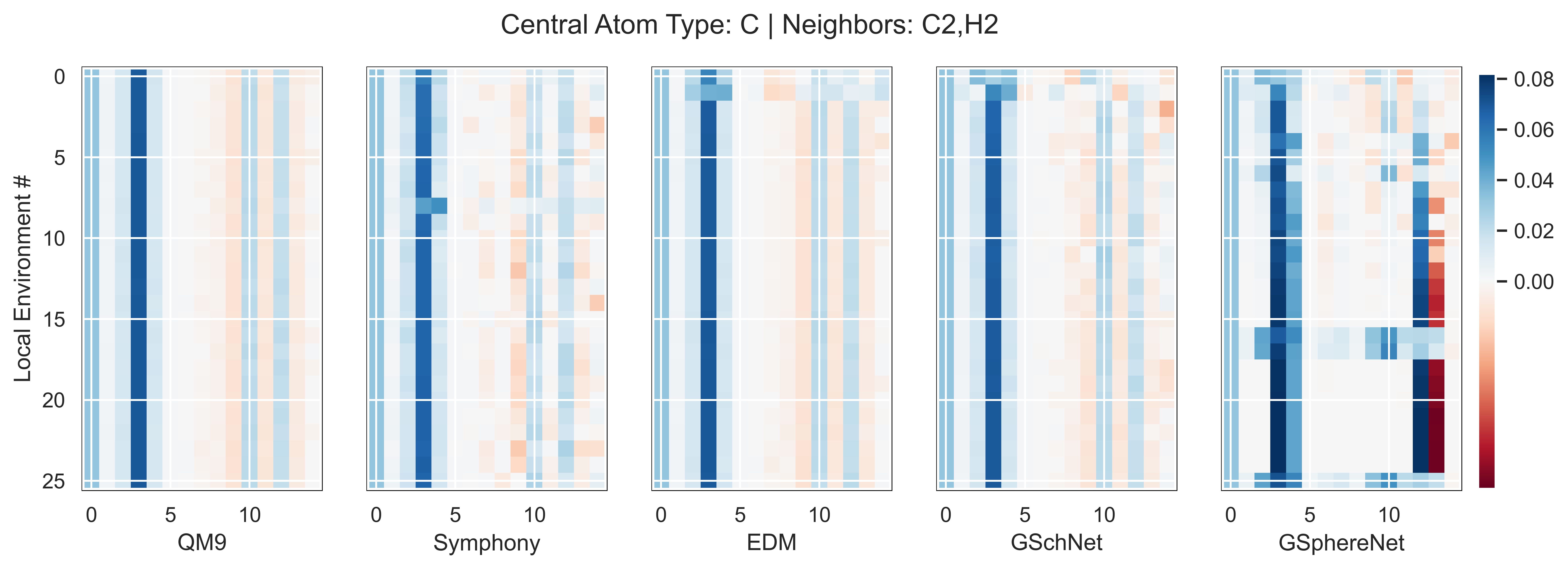}
    \includegraphics[width=\textwidth]{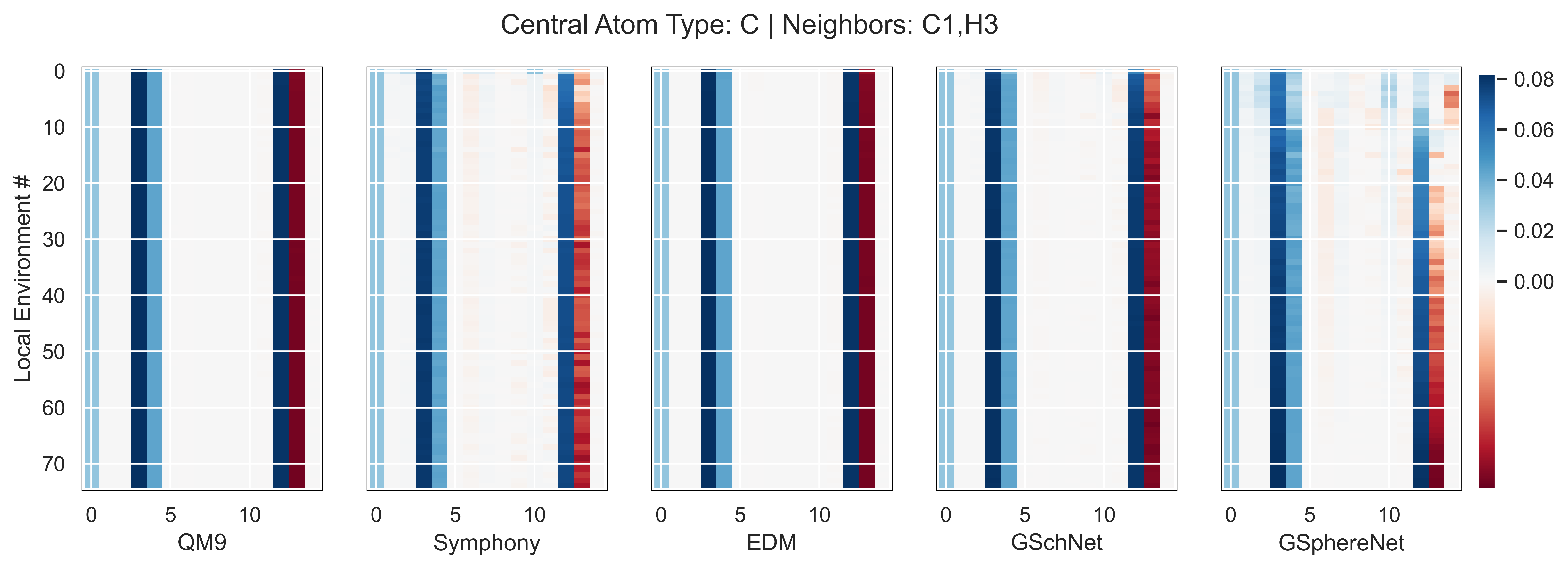}
    \caption{Bispectra of local environments of type C: C2,H2 and type C: C1,H3 respectively. Each row corresponds to a sample of the bispectrum (an array of length 15). Every entry of the bispectra is colored by value according to the colorbar on the right. }
    \label{fig:bispectra}
\end{figure}

As seen in \autoref{fig:bispectra}, we see that Symphony's sampled bispectra (second from left) have a slightly different distribution relative to those from QM9 in the two most frequent local environments.

\subsection{Bond Lengths in Sampled Molecules}
\begin{figure}[p]
    \centering
    \includegraphics[width=\textwidth]{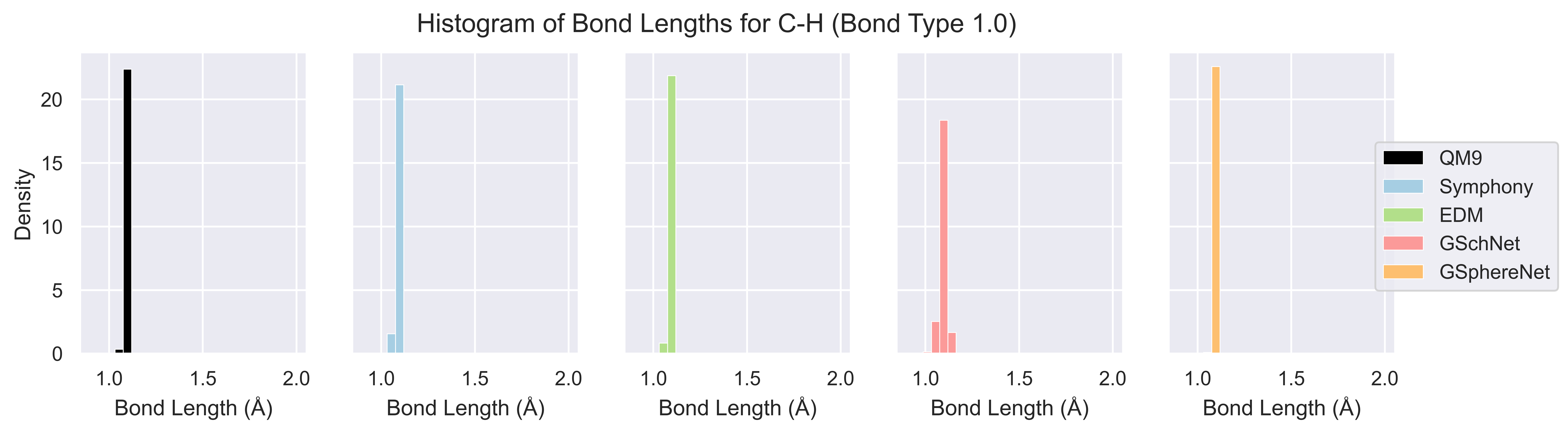}
    \includegraphics[width=\textwidth]{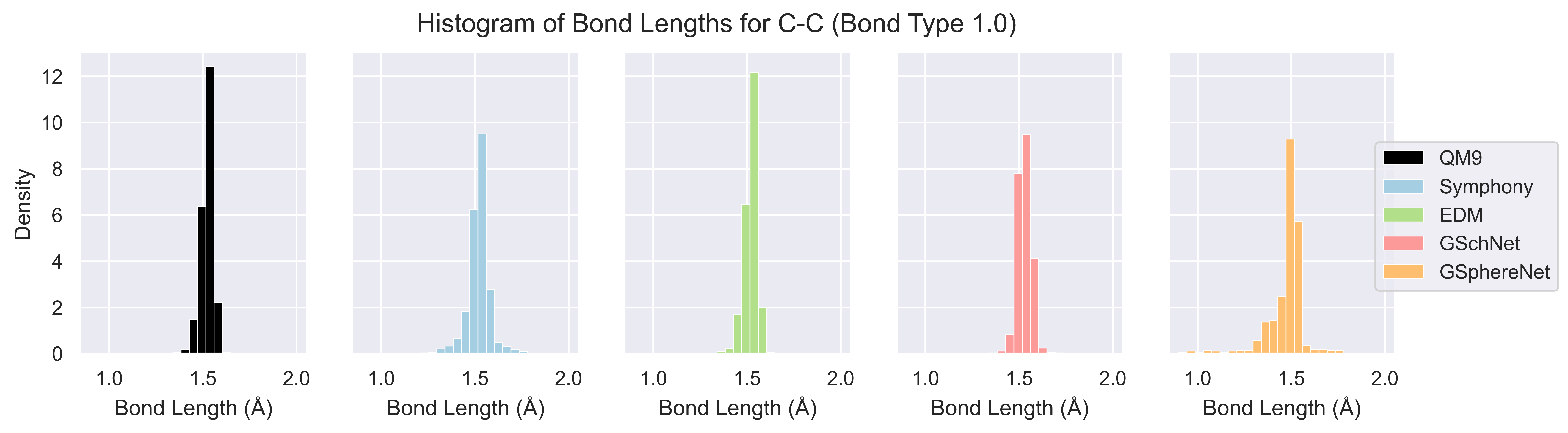}
    \includegraphics[width=\textwidth]{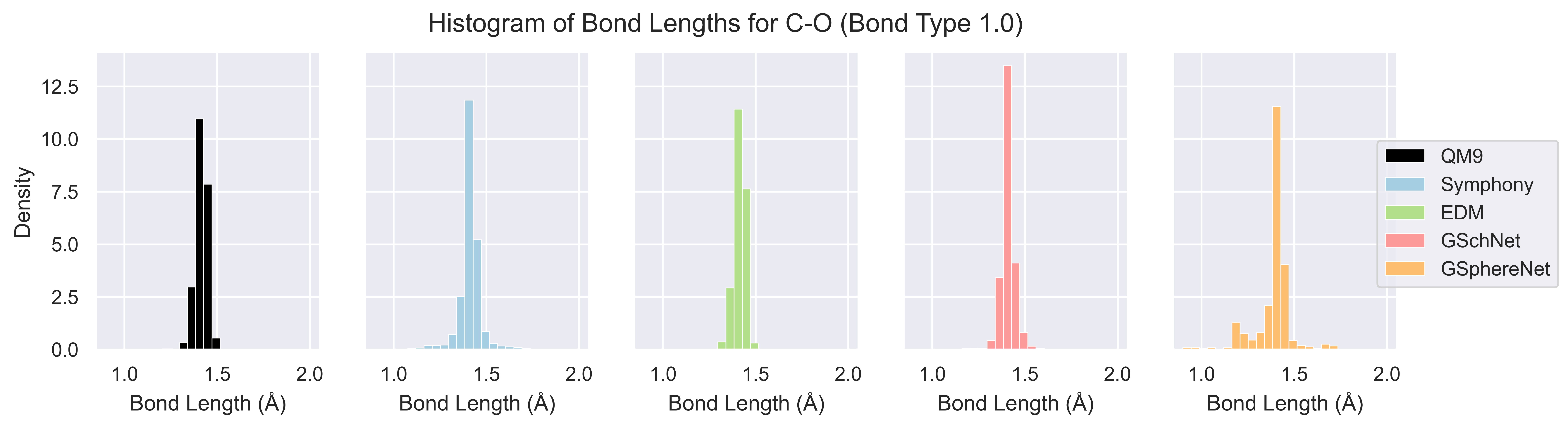}
    \includegraphics[width=\textwidth]{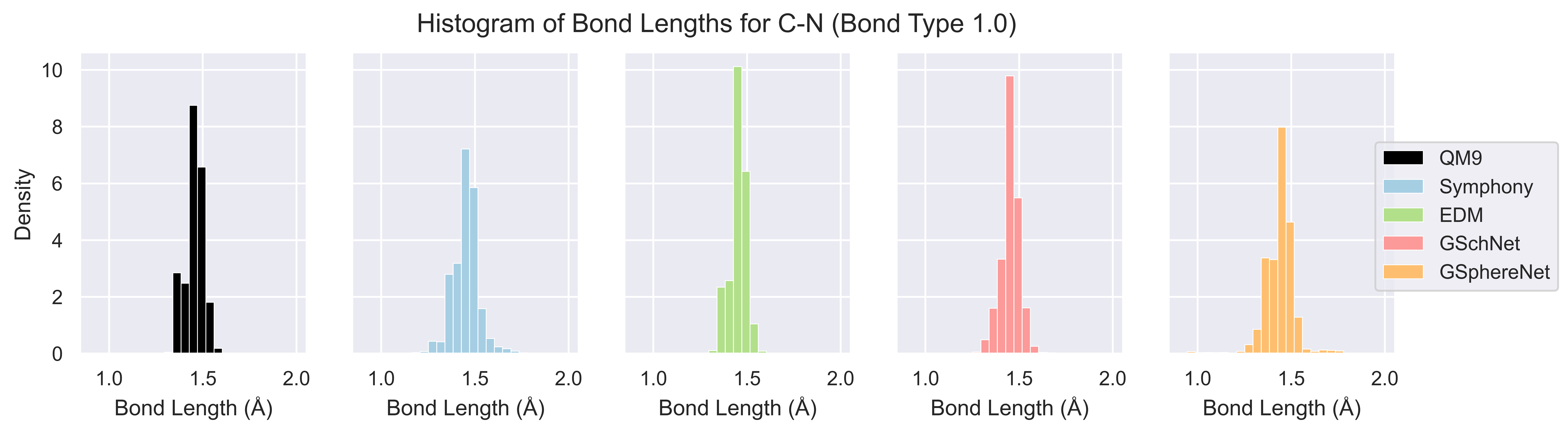}
    \includegraphics[width=\textwidth]{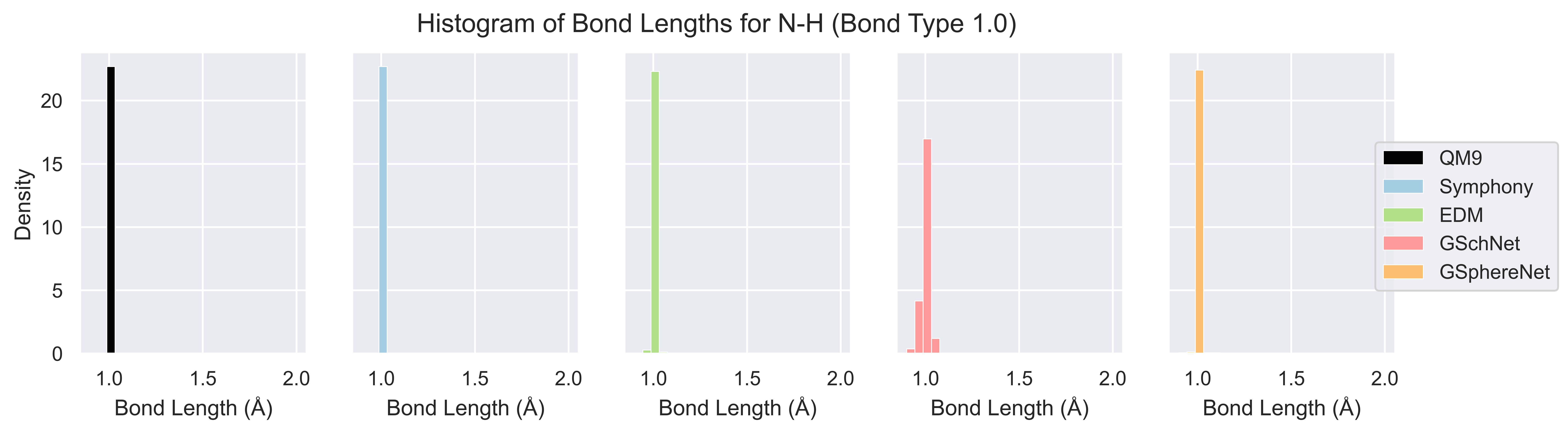}
    \caption{Histogram of bond lengths for the five most frequent bonds in QM9.}
    \label{fig:bond_lengths_1}
\end{figure}

\begin{figure}[p]
    \includegraphics[width=\textwidth]{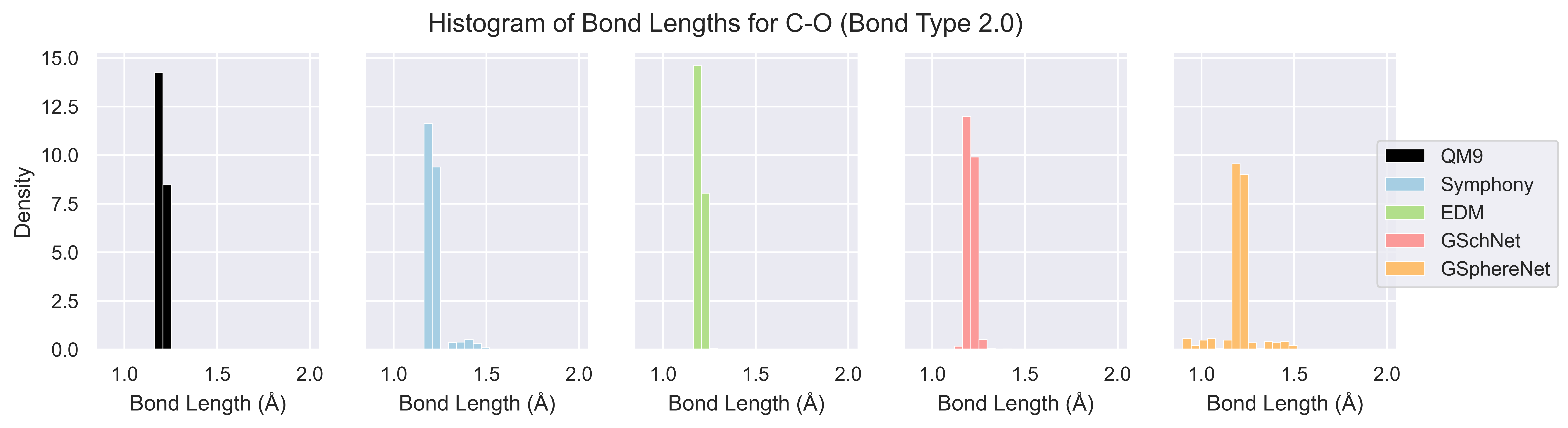}
    \includegraphics[width=\textwidth]{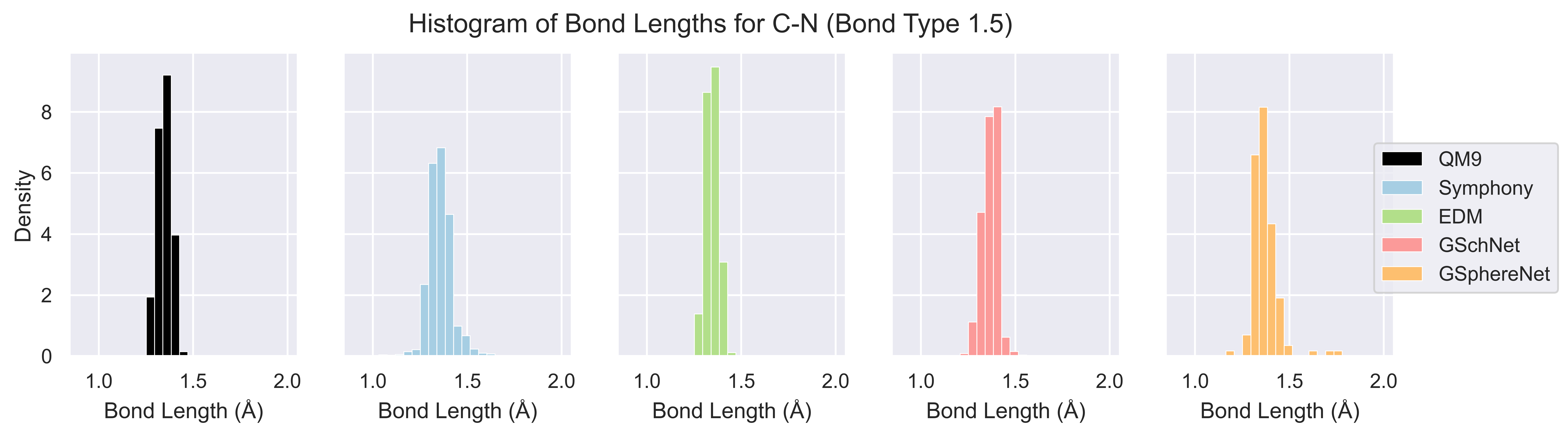}
    \includegraphics[width=\textwidth]{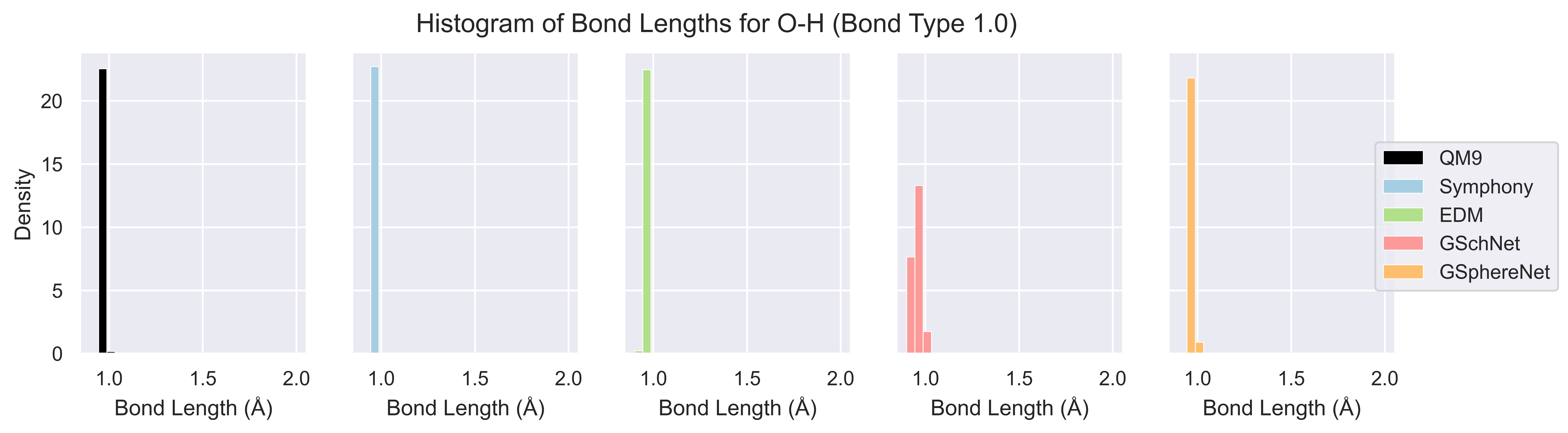}
    \includegraphics[width=\textwidth]{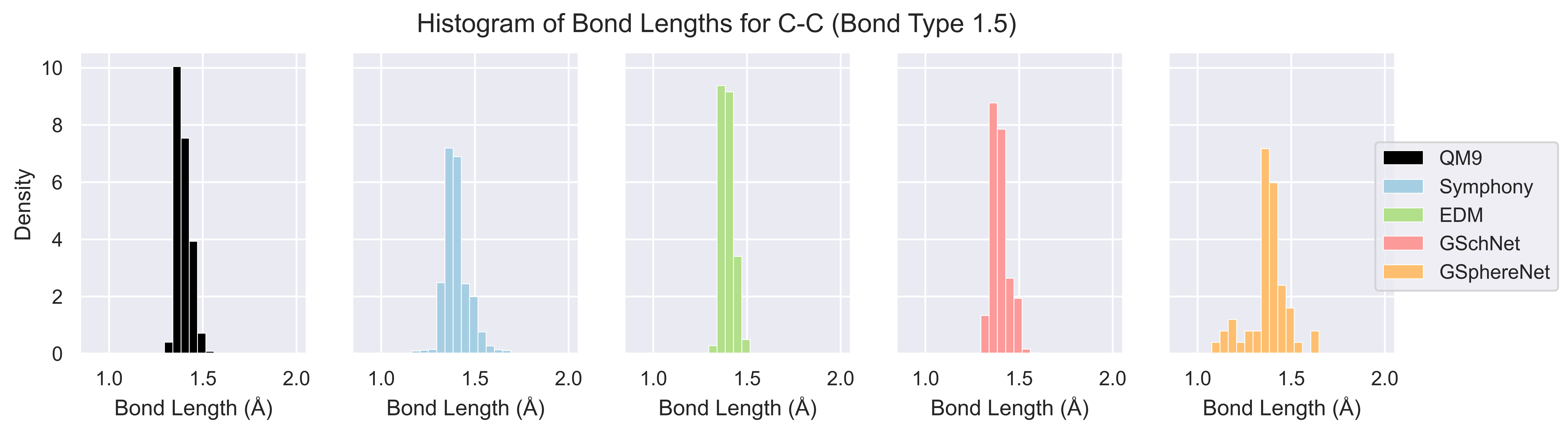}
    \includegraphics[width=\textwidth]{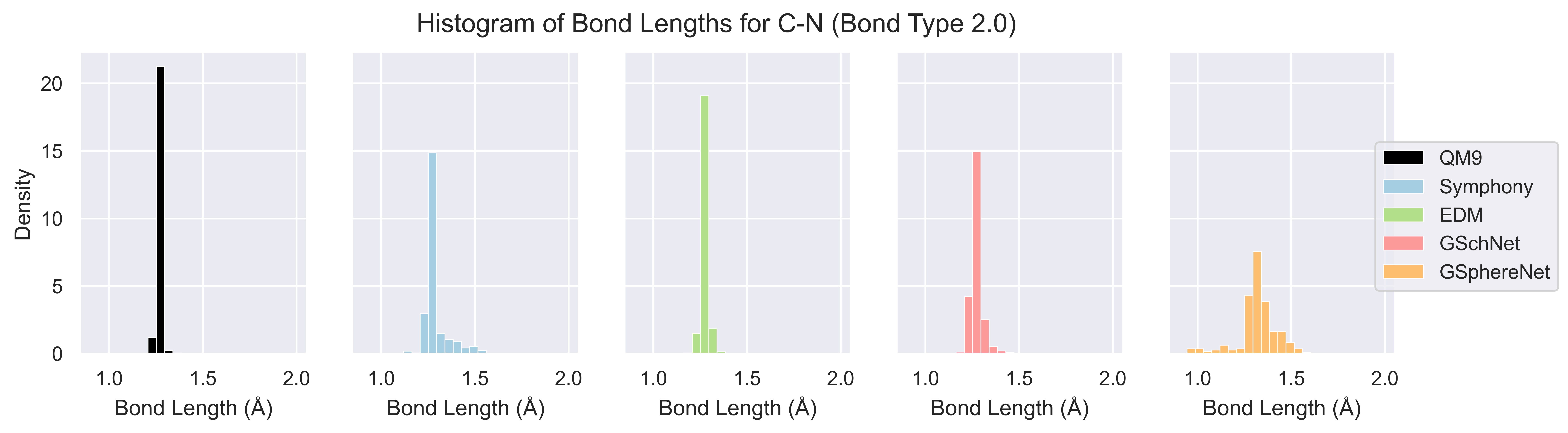}
    \caption{Histogram of bond lengths for the sixth to tenth most frequent bonds in QM9.}
    \label{fig:bond_lengths_2}
\end{figure}

From \autoref{fig:bond_lengths_1} and \autoref{fig:bond_lengths_2}, we see that Symphony's bond length distribution tends to be wider than those of QM9, hurting its MMD score relative to EDM. Improving this aspect is an ongoing effort; but we believe that the bond lengths are still quite reasonable.


\subsection{Atom Type Counts}
\begin{figure}[p]
    \centering
    \includegraphics[width=\textwidth]{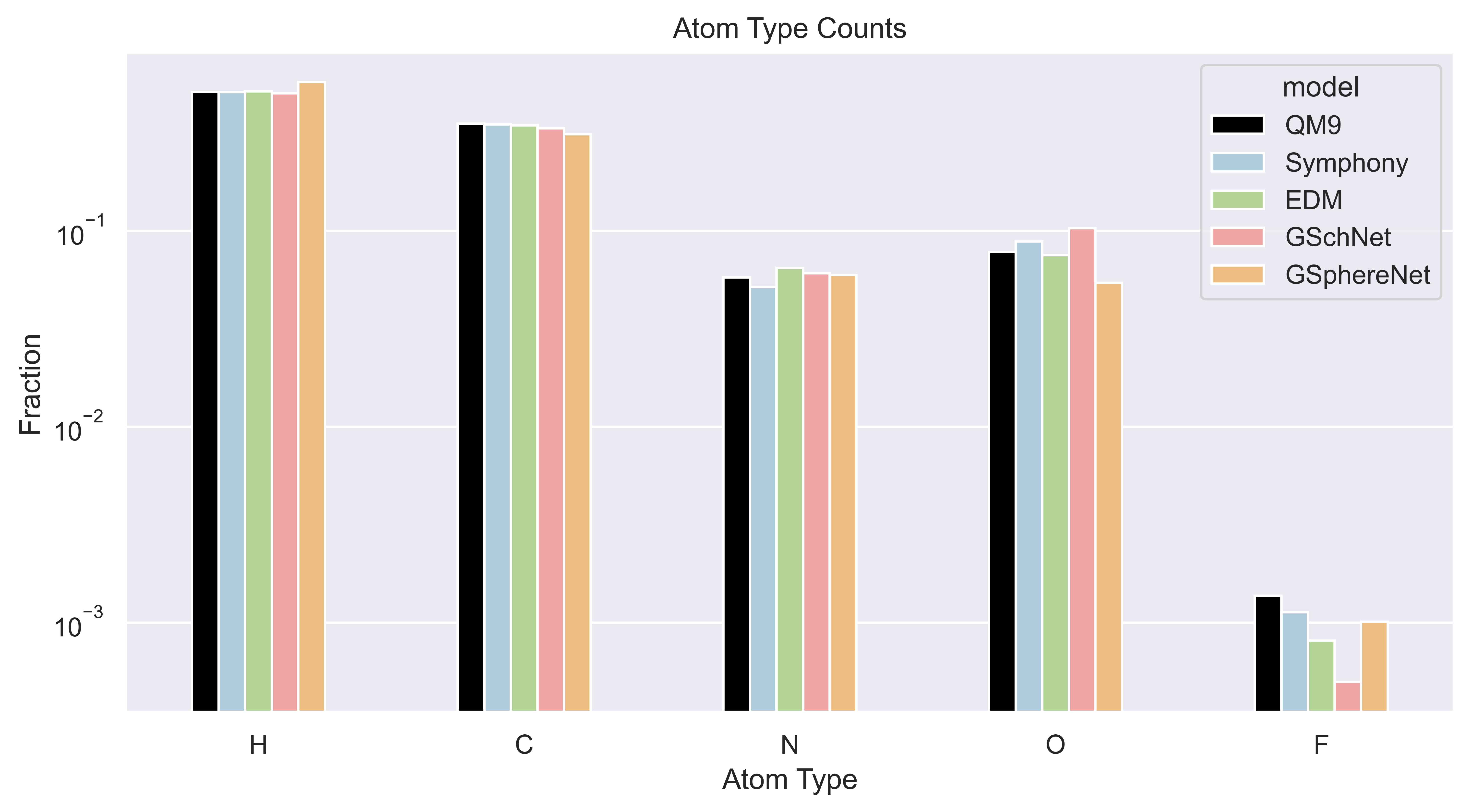}
    \caption{Frequency of atom type counts in generated molecules on a log-scale.}
    \label{fig:atom_type_counts}
\end{figure}

As seen in \autoref{fig:atom_type_counts}, all models are able to reasonably capture the distribution of atom types in QM9; Symphony performs especially well here.

\subsection{Ring Sizes}
\begin{figure}[p]
    \centering
    \includegraphics[width=\textwidth]{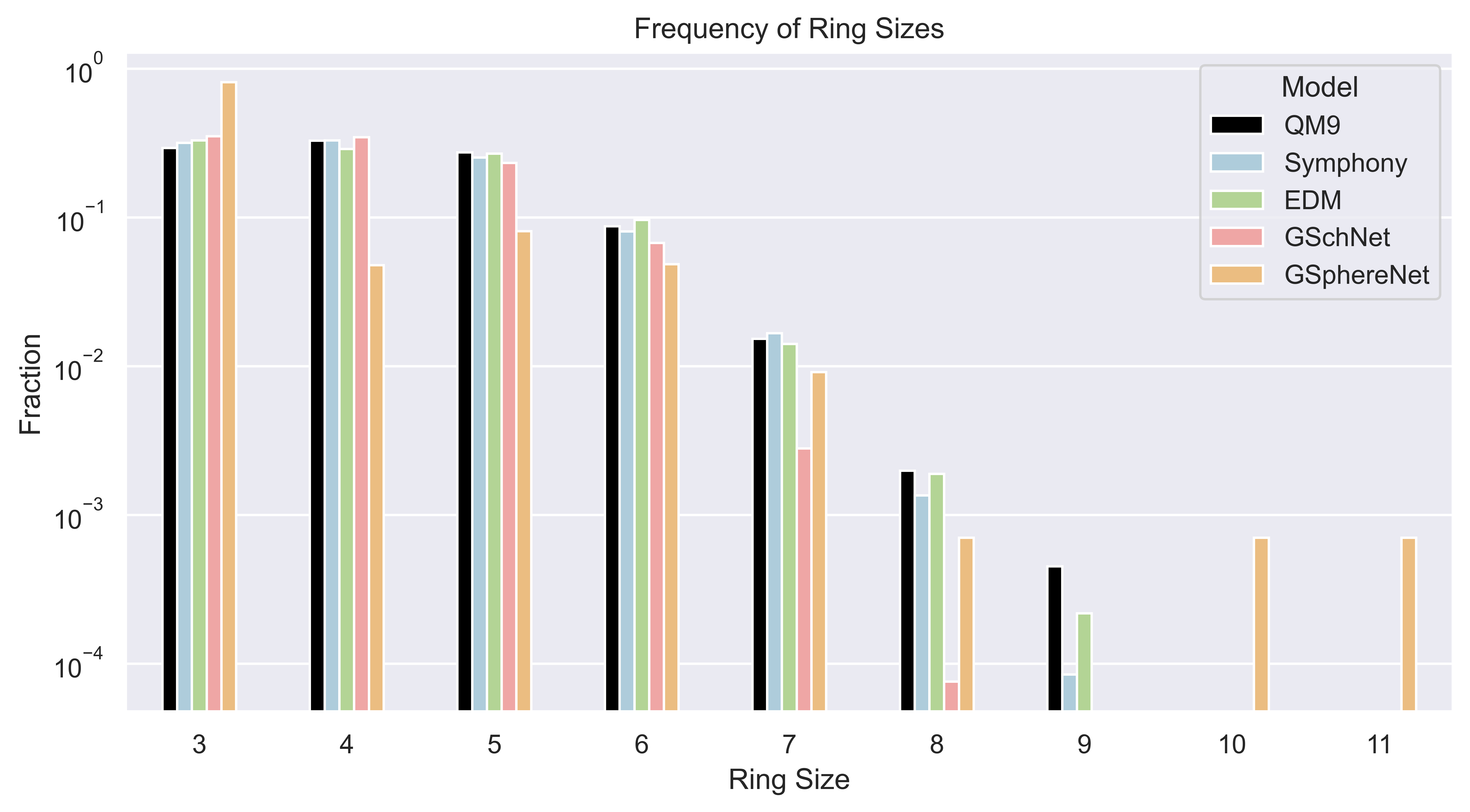}
    \caption{Frequency of ring sizes in generated molecules on a log-scale.}
    \label{fig:ring_sizes}
\end{figure}

We also extracted all rings using RDKit \citep{rdkit} and counted their relative frequency, in \autoref{fig:ring_sizes}. G-SphereNet seems to produce either very large or very small rings. The other models seem to capture the distribution of ring sizes well.

\section{Proof of E(3)-Equivariance}
\label{sec:equivariance-proof}

\begin{theorem}
\label{thm:equivariance-proof}
\upshape
Suppose \textsc{Embedder} produces $O(3)$-equivariant and translation-invariant features $h_{v,l} = \textsc{Embedder}(\gS^n)_{v,l}$ for every atom $v$. Then, $p^\text{position}$ is $O(3)$-equivariant and translation-invariant (and hence, $E(3)$-equivariant):
\begin{align*}
    p^\text{position}(\R \r_{n + 1}  + \T \ | \ f_{n + 1}, Z_{n + 1}; \R \gS^{n} + \T)
    = p^\text{position}(\r_{n + 1} \ | \ f_{n + 1}, Z_{n + 1}; \gS^{n})
\end{align*}
\end{theorem}

\textbf{Proof}: We first show that $p^\text{position}$ is $O(3)$-equivariant. We have:
\begin{align*}
    \textsc{Embedder}(\R\gS^n)_{v,l} = D^l(\R)^T \textsc{Embedder}(\gS^n)_{v,l}
\end{align*}
for every atom $v$ and degree $l$. Note that because $Z_{n + 1}$ is rotationally invariant, it immediately follows from \autoref{eqn:position_coeffs} and the above, that $c_l$ is also $E(3)$-equivariant with degree $l$:
\begin{align*}
    c_l(r; \R\gS^{n}, f_{n + 1}, Z_{n + 1}) =  c_l(r; \gS^{n}, f_{n + 1}, Z_{n + 1})
\end{align*}
Now, as the Wigner D-matrices are always unitary, we have:
\begin{align*}
f^\text{position}(\R\r; \R\gS^{n}, f_{n + 1}, Z_{n + 1})
&= \sum_{l = 0}^\infty c_l(r; \R\gS^{n}, f_{n + 1}, Z_{n + 1})^T Y_l(\R\hat{\textbf{r}}_{ij}) \\
&= \sum_{l = 0}^\infty c_l(r; \gS^{n}, f_{n + 1}, Z_{n + 1})^T D^l(\R) D^l(\R)^T Y_l(\hat{\textbf{r}}_{ij}) \\
&= \sum_{l = 0}^\infty c_l(r; \gS^{n}, f_{n + 1}, Z_{n + 1})^T Y_l(\hat{\textbf{r}}_{ij}) \\
&= f^\text{position}(\r; \gS^{n})
\end{align*}
by definition. Thus, we are guaranteed that $f^\text{position}$ is $O(3)$-equivariant.
Note that applying a pointwise non-linearity ($\exp$) to $f^\text{position}$ and a rotationally invariant normalization does not change $O(3)$-equivariance. Thus, $p^\text{position}$ is $O(3)$-equivariant as well.

For translations, note that $p^\text{position}$ is described relative to the focus atom $f_{n + 1}$.
Thus, as \textsc{Embedder} is translation-invariant:
\begin{align*}
    \textsc{Embedder}(\gS^n + \T)_{v,l} = \textsc{Embedder}(\gS^n)_{v,l}
\end{align*}
$p^\text{position}$ will be translation-equivariant:
\begin{align*}
p^\text{position}(\r_{n + 1} + \T \ | \ f_{n + 1}, Z_{n + 1}; \gS^{n} + \T) = p^\text{position}(\r_{n + 1}  \ | \ f_{n + 1}, Z_{n + 1}; \gS^{n})
\end{align*}
In conclusion, $p^\text{position}$ is $O(3)$-equivariant and translation-equivariant, and hence $E(3)$-equivariant. Thus, \hyperref[prop:2]{\textbf{Property (2)}} is satisfied. {$\hfill\blacksquare$}

\begin{theorem}
\label{thm:permutation-equivariance-proof}
\upshape
Suppose \textsc{Embedder} produces permutation-equivariant features $h_{v,l} = \textsc{Embedder}(\gS^n)_{v,l}$ for every atom $v$. Then, $p^\text{focus}$ is permutation-equivariant, while $p^\text{species}$ and $p^\text{position}$ are permutation-invariant:
\begin{align*}
        p^\text{focus}(\pi(f_{n + 1}); \pi \gS^{n}) &= p^\text{focus}(f_{n + 1}; \gS^{n}) \\
        p^\text{species}(Z_{n + 1} \ | \ \pi(f_{n + 1}); \pi\gS^{n}) &= p^\text{species}(Z_{n + 1} \ | \ f_{n + 1}; \gS^{n})
        \\
        p^\text{position}(\r_{n + 1} \ | \ \pi(f_{n + 1}), Z_{n + 1}; \pi\gS^{n})
    &= p^\text{position}(\r_{n + 1} \ | \ f_{n + 1}, Z_{n + 1}; \gS^{n})
\end{align*}
where $\pi$ represents a permutation of the atoms of $\gS^n$.

\textbf{Proof}: 
Because $\textsc{Embedder}$ is permutation-equivariant:
\begin{align*}
    \textsc{Embedder}(\pi\gS^n)_{\pi(v),l} = \textsc{Embedder}(\gS^n)_{v,l}
\end{align*}
for each atom $v$.
Then, from \autoref{eqn:focus_and_atom_type}: 
\begin{align*}
    p^\text{focus}(\pi(f_{n + 1}) ; \pi \gS^n)  &= \textsc{MLP}(\textsc{Embedder}(\pi \gS^n)_{\pi(f_{n + 1}), 0}) \\
    &= \textsc{MLP}(\textsc{Embedder}(\gS^n)_{f_{n + 1},0}) \\
    &= p^\text{focus}(f_{n + 1}) ; \gS^n)
\end{align*}
as claimed.
Similarly, 
\begin{align*}
    p^\text{species}(Z_{n + 1} \ | \ \pi(f_{n + 1}); \pi\gS^{n})  &= \textsc{MLP}(\textsc{EmbedAtomType}(Z_{n + 1}) \cdot \textsc{Embedder}(\pi\gS^n)_{\pi(f_{n + 1}),0}) \\
    &= \textsc{MLP}(\textsc{EmbedAtomType}(Z_{n + 1}) \cdot \textsc{Embedder}(\gS^n)_{f_{n + 1},0}) \\
    &=
    p^\text{species}(Z_{n + 1} \ | \ f_{n + 1}; \gS^{n})
\end{align*}
For $p^\text{position}$, it is sufficient to show that the coefficients $c_l(r)$ are permutation-equivariant:
\begin{align*}    
    c_l(r;  \pi(f_{n + 1}), Z_{n + 1}, \pi \gS^{n})
    &= \textsc{Linear}(\textsc{Embedder}(\pi\gS^n)_{\pi(f_{n + 1}),l} \otimes \textsc{EmbedAtomType}(Z_{n + 1})) \\
    &= \textsc{Linear}(\textsc{Embedder}(\gS^n)_{f_{n + 1},l} \otimes \textsc{EmbedAtomType}(Z_{n + 1})) \\
    &= c_l(r;  f_{n + 1}, Z_{n + 1}, \gS^{n})
\end{align*}
Thus, all distributions transform as expected. {$\hfill\blacksquare$}

\end{theorem}
\section{Details of Models}
\label{sec:model-details}

\subsection{Embedders}
Here, we describe E3SchNet and NequIP \citep{nequip} which we use to embed the atoms in each fragment into $E(3)$-equivariant features. As shown in \autoref{sec:equivariance-proof}, we require these models to be $E(3)$-equivariant. 

Both of these models are geometric message-passing neural networks, a type of graph neural network \citep{sanchez-lengeling2021a,daigavane2021understanding} that respects the symmetries of 3D structures.
In particular, E3SchNet as the $\textsc{Embedder}$ for the focus and atom type prediction, and NequIP as the $\textsc{Embedder}$ for the position prediction. Unlike previous autoregressive models which utilized a shared embedder for all tasks, we found that using different embedders for these two tasks performed much better in our experiments.

Given the fragment $\gS^{n}$, we define the neighbour of each atom $i \in \gS^{n}$ by a Euclidean distance cutoff $\leq d_\text{max}$:
\begin{align}
    \label{eqn:neighbor}
    \gN(i) = \{ j \in \gS^{n} \ |  \ \norm{\r_{ij}} \leq d_\text{max} \} 
\end{align}
Initially, the features $h_i^{(0)}$ of each atom $i$ in $\gS^{n}$ are set as the embedding of its atomic number $Z_i$.
At each iteration $t$, the features $h_i^{(t)}$ is updated using the atom's features $h_i^{(t - 1)}$ and its neighbour's features $h_j^{(t - 1)}$ where $j \in \gN(i)$ from the previous round. The final embedding for atom $i$ is returned as $h_i^{(T)}$ where $T$ is the number of message-passing iterations.
\autoref{alg:mpnn} formally shows the operations of a general message passing neural network.

\begin{algorithm}
\caption{General Operation of a Message Passing Neural Network}
\label{alg:mpnn}
\begin{algorithmic}
\Require Fragment $\gS^n$, Message Passing Iterations $T$, Cutoff $d_{\text{max}}$, Update Function \textsc{Update}
\State Compute neighbor lists for each atom in $\gS^n$ according to \autoref{eqn:neighbor}.
\For{$i = 1, 2, \ldots, n$}:
\Let{$h_i^{(0)}$}{$\textsc{ScalarEmbedding}
(Z_i)$}
\EndFor
\For{$t = 1, 2, \ldots, T$}:
\For{$i = 1, 2, \ldots, n$}:
\Let{$h_{\gN(i)}^{(t - 1)}$}{$\{h_j^{(t - 1)} \ | \ j \in \gN(i)\}$}
\Let{$h_i^{(t)}$}{$\textsc{Update}(h_i^{(t - 1)}, h_{\gN(i)}^{(t - 1)})$}
\EndFor
\EndFor
\State \Return $\{h_i^{(T)}\}_{i = 1}^n$  
\end{algorithmic}
\end{algorithm}

Different message-passing networks differ in their choice of \textsc{Update} function.
Following \citet{nequip}, the \textsc{Update} for NequIP is defined as:
$$
\textsc{Update}(h_i^{(t - 1)}, h_{\gN(i)}^{(t - 1)})
= h_i^{(t - 1)} + \frac{1}{C}\sum_{j \in \gN(i)} \sum_{l = 0}^{l_\text{max}} R_\Theta(\norm{\r_{ij}}) Y^l(\hat{\textbf{r}}_{ij}) \otimes h_j^{(t - 1)} 
$$
$R_\Theta(\cdot)$ is a learned multi-layer perceptron (MLP). We set $C = 20, d_\text{max} = 5\text{A}, \lmax = 5$, and $T = 3$ here. For clarity, we assume the decomposition of the tensor product into a direct sum of irreducible representations of $O(3)$ above.

E3SchNet is our generalization of the SchNet model \citep{schnet} that was used in \citep{gschnet} to produce higher-degree $E(3)$-equivariant features. The \textsc{Update} function for E3SchNet is defined as:
$$
\textsc{Update}(h_i^{(t - 1)}, h_{\gN(i)}^{(t - 1)}) =
h_i^{(t - 1)} + 
\textsc{Linear}\left(\sum_{j \in \gN(i)}  \sum_{l = 0}^{l_\text{max}} W_{ijl} \cdot \left(h_j^{(t - 1)} \otimes Y^l(\hat{\textbf{r}}_{ij})\right)  \right)
$$
where $W_{ijl}$ are scalars computed via:
\begin{align*}
    W_{ijl} = 
    \textsc{Linear}(\sigma(\textsc{Cutoff}(\norm{\r_{ij}}) \cdot \textsc{RadialBasis}(\norm{\r_{ij}})))
\end{align*}
We use the Gaussian radial basis functions, following SchNet. In fact, for $\lmax = 0$, E3SchNet reduces exactly to the standard SchNet. We set $\lmax = 2$, as we find that the benefits of using even higher degree features for the focus and atom type prediction task are minimal. The cutoff is again $5$A.

We see that NequIP and E3SchNet guarantee permutation-equivariance, translation invariance and $O(3)$-equivariance, and hence satisfy the requirements for \textsc{Embedder} in \autoref{sec:equivariance-proof}.

We implement Symphony with the \texttt{e3nn-jax} library that utilizes the JAX \citep{jax2018github} framework for creating efficient $E(3)$-equivariant machine learning models.

\subsection{Training Details}
\label{sec:training-details}
We set $\sigma_{\text{true}}^2 = 10^{-5}$ and express the Dirac delta distribution in the spherical harmonic basis upto $\lmax = 5$, as explained in \autoref{sec:delta-functions}.
The predicted distributions $p^\text{focus}, p^\text{species}$ 
and $p^\text{position}$ are learned by minimizing the KL divergence to their true counterparts. We found that adding a small amount of zero-centered Gaussian noise $\sigma^2 = 2.5 \times 10^{-3}$ to all input atom positions helped with robustness. All parameters in the \textsc{Embedder}, \textsc{MLP} and \textsc{Linear} layers are trained with the Adam \citep{adam} optimizer with a learning rate of $5 \times 10^{-4}$. We chose the parameters that achieved the lowest loss on the validation set over $8000000$ training steps with a batch size of $16$ fragments.

\subsection{Data Details}
\label{sec:data-details}
Following EDM \citep{edm},
we obtained the QM9 \citep{qm9} dataset using the DeepChem library \citep{deepchem},
and filtered out $3054$ ‘uncharacterized’ molecules (available at
\url{https://springernature.figshare.com/ndownloader/files/3195404}) which rearranged significantly
during geometry optimization, giving us exactly $130831$ molecules. Symphony was trained used the same splits as EDM:
$100000$ molecules to train, $13083$ molecules for validation and $17748$ molecules for test, obtained from a random permutation of the molecules.

\subsection{Baseline Model Details}
\label{sec:baseline-details}
For the baseline models, we used the pretrained EDM model at 
\url{https://github.com/ehoogeboom/e3_diffusion_for_molecules} and the pretrained G-SphereNet model
at \url{https://github.com/divelab/DIG/tree/dig-stable/examples/ggraph3D/G_SphereNet}. We retrained
the G-SchNet model on the EDM splits following
\url{https://github.com/atomistic-machine-learning/G-SchNet}. The samples (in \texttt{.xyz} format)
of all models used for evaluation is available at this URL: \url{https://figshare.com/s/a17ccface17f0c22f15a}.

\section{Learning and Sampling from Position Distributions}
\label{sec:position-distributions}

In this section, we drop the superscript from $p^\text{position}$ as it should be clear from context.

\subsection{Training}
To recap \autoref{sec:symphony-design}, Symphony predicts coefficients $c^{\text{ch}}_{l}(r;  \gS^{n})$ to represent the position distribution $p$:
\begin{align*}
f(r, \theta, \phi) &= \log \sum_{\text{channel ch}} \exp \sum_{l = 0}^{\infty} c^{\text{ch}}_{l}(r; \gS^{n})^T Y_l(\theta, \phi) \\
p(r, \theta, \phi) &= \frac{1}{Z}
 \exp{f(r, \theta, \phi)}
\end{align*}
where $Z$ is the partition function.

As mentioned in \autoref{sec:training-details}, 
the coefficients are learned by minimizing the KL divergence to the target distribution $q$:
\begin{align*}
KL(q \ || \ p) = \int_\Omega q(\r) \log \frac{q(\r)} {p(\r)} d\r =  \int_\Omega q(\r) \log q(\r) d\r - \int_\Omega q(\r) f(\r) d\r + \log Z
\end{align*}
Following the notation of \autoref{sec:symphony-design}, $\Omega$ represents the set  $\{ r \in [0, \infty),
\theta \in [0, \pi], \phi \in [0, 2\pi)\}$ which is all space in spherical coordinates.

For training, we only need the unnormalized logits $f$ and not the normalized distribution $p$. This is identical to the log-sum-exp trick when training with cross-entropy loss for a classification problem. Unlike the classification case where the number of classes is finite, the integral above must be computed over all of $r$, $\theta$ and $\phi$ which is an infinite set. To numerically approximate this integral, we use a uniform grid on $r$ and a Spherical Gauss-Legendre quadrature on the sphere at each value of $r$. As discussed in \autoref{sec:symphony-design}, the uniform grid on $r$ spans $64$ values from $0.9$A to $2.0$A which is more than sufficient to cover all bond lengths in organic molecules. The Spherical Gauss-Legendre quadrature is a product of two quadratures: a 1D Gauss-Legendre quadrature with $180$ points over $\cos \theta \in [-1, 1]$, and a uniform grid of $359$ points over $[0, 2\pi)$ for $\phi$.

Symphony predicts the coefficients $c_l(r)$ of $f$ which can be used to evaluate $f(r, \theta, \phi)$ at any point. This evaluation for a spherical grid of $(\theta, \phi)$ values can be done quickly via a Fast Fourier Transform (FFT) that is implemented in \texttt{e3nn-jax}. We perform this FFT procedure for each sphere defined by a radial grid point $r$.

\subsection{Sampling}
Once the model is learnt, we need to sample from the distribution $p$. A key advantage of predicting the coefficients $c_l(r)$ of $f_\theta(r, \theta, \phi)$ is that a different resolution of angular grid can be chosen for sampling than that of training. 
We simply evaluate $f(r, \theta, \phi)$ on the quadrature grid as before, apply the exponential, and normalize via numerical integration to get $p(r, \theta, \phi)$. We first marginalize over $\theta, \phi$ to obtain a distribution $p(r)$ to sample a radius $r$. Then, we sample one of the angular grid points $(\theta, \phi)$ for the sphere corresponding to this radius $r$. Overall, this procedure gives us a sample from $p(r, \theta, \phi)$.

In \autoref{sec:validity_vs_resolution}, we assess how the validity of molecules generated by Symphony varies as the grid resolution is varied.

Note that our sampling procedure is much simpler than that of \citet{simm2021symmetryaware}, which uses rejection sampling with a uniform base distribution. We perform some quantitative experiments with the parametrization of \citet{simm2021symmetryaware} in \autoref{sec:learning-random-signals}.

While we are primarily interested in 
learning distributions over $\mathbb{R}^3$
which are equivariant under $E(3)$,
there has been prior work in learning distributions over manifolds
\citet{cohen2015harmonic, murphy2022implicitpdf},
where the issue of
estimating the partition function are also
solved by discretizing over an appropriate
domain.

\section{Details of Metrics}
\label{sec:metrics-details}

\subsection{PoseBusters}
\label{sec:posebusters-details}
\autoref{tab:posebusters_description} provides details of the Posebusters tests used in \autoref{tab:posebusters_metrics}. We use the default parameters from their framework.

\begin{table}[h]
    \centering
    \begin{tabularx}{\textwidth}{cX}
        \toprule
        Test & Description \\
        \midrule
        All Atoms Connected & There exists a path along bonds between any two atoms in the molecule. \\
        Reasonable Bond Lengths  &  The bond lengths in the input molecule are within $0.75$ of the lower and $1.25$ of the upper bounds determined by distance geometry. \\
        Reasonable Bond Angles  & The angles in the input molecule are within $0.75$ of the lower and $1.25$ of the upper bounds determined by distance geometry. \\
        Aromatic Rings Flatness & All atoms in aromatic rings with $5$ or $6$ members are within $0.25$A of the closest shared plane. \\
        Double Bonds Flatness & The two carbons of aliphatic carbon-carbon double bonds and their four neighbours are within $0.25$A of the closest shared plane. \\
        Reasonable Molecule Energy & The calculated energy of the input molecule is no more than $100$ times the average energy of an ensemble of $50$ conformations generated for the input molecule. The energy is calculated using the UFF \citep{uff} in RDKit \citep{rdkit} and the conformations are generated with ETKDGv3 \citep{etkdg} followed by force field relaxation using the UFF with up to $200$ iterations. \\
        No Internal Steric Clash & The interatomic distance between pairs of non-covalently bound atoms is above $0.8$ of the lower bound determined by distance geometry. \\
    \end{tabularx}
    \caption{Description of each intramolecular PoseBusters test, taken from Table 4 of \citet{posebusters}.}
    \label{tab:posebusters_description}
\end{table}

\subsection{Maximum Mean Discrepancy}
\label{sec:mmd-details}
The Maximum Mean Discrepancy (MMD), introduced in \citet{mmd}, measures how different two distributions $p_X$ and $p_Y$ are, given a kernel function $k$. Formally, the MMD is defined as:
\begin{align*}
    \text{MMD}(p_X, p_Y) = \sqrt{\underset{X, X' \sim p_X}{\E}[k(X, X')] + \underset{Y, Y' \sim p_X}{\E}[k(Y, Y')] - \underset{X \sim p_X, Y \sim p_Y}{\E}[k(X, Y)]}
\end{align*}
From the above equation, we see that the MMD can be easily estimated with samples from each distribution. We choose $k$ as the sum of Gaussian kernels at different scales:
\begin{align*}
    k(X, X') = \sum_{i = 0}^{29}\exp(-10^{\left(\frac{i}{5} - 3\right)} \cdot \norm{X - X'}^2)
\end{align*}

\section{The Advantage of Using Multiple Channels of Spherical Harmonics}
\label{sec:channels-example}

\subsection{An Example with the Octahedron}
\begin{figure*}[!h]
    \centering
    \includegraphics[width=0.65\textwidth]{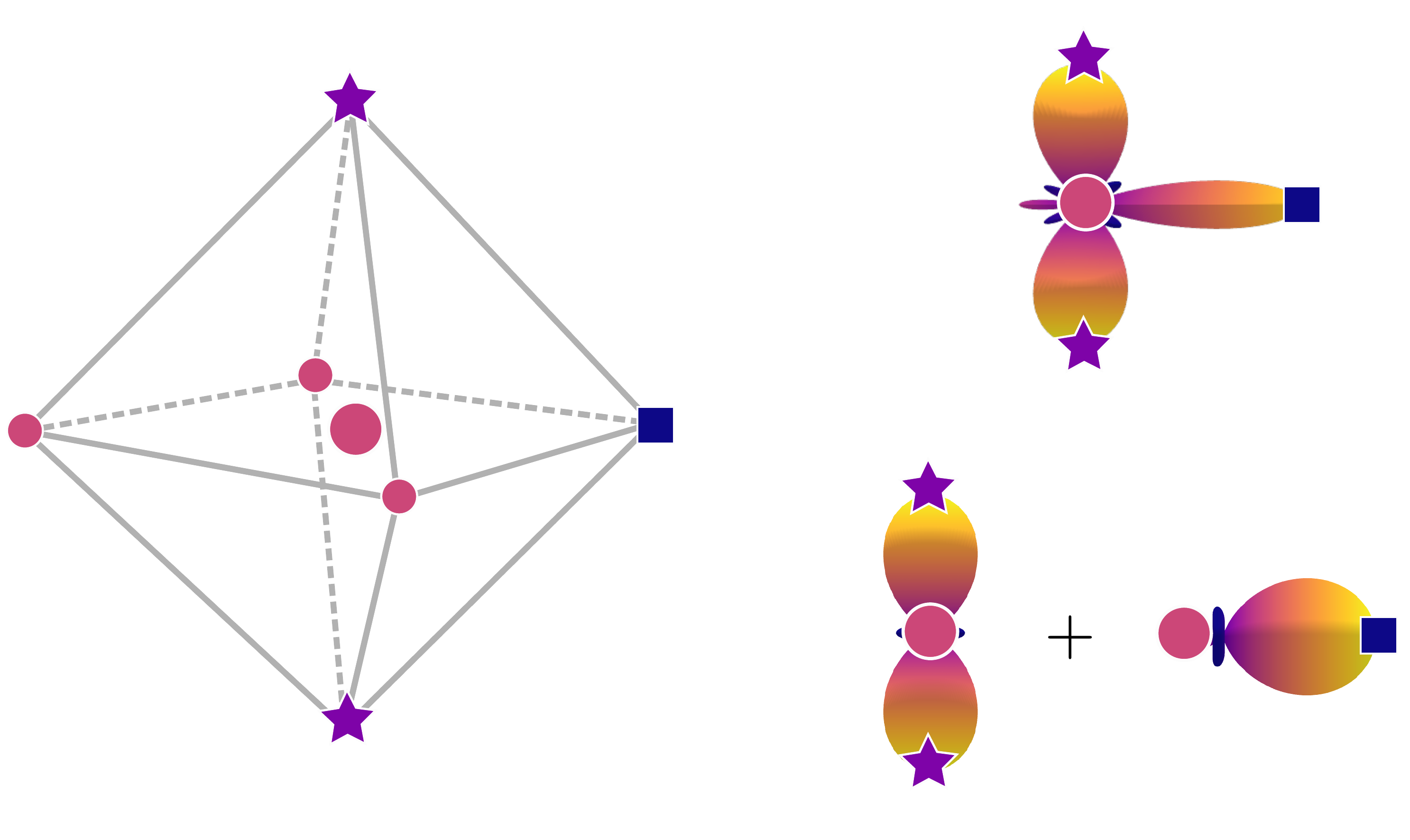}
    \caption{Usually, we would require $\lmax = 4$ to represent $p^{\text{position}}$ for
    the `stars' and `square' atoms, centered at the red central atom. With two channels, we
    only need up to $\lmax = 2$ each. 
    }
    \label{fig:octahedron_channels}
\end{figure*}

\autoref{fig:octahedron_channels} shows how adding a second channel helps reduce the 
effective $\lmax$ needed to represent $p^{\text{position}}$. The atoms depicted by red 
circles have been placed already, and the atom at the center of the octahedron has been 
chosen as the focus. To accurately capture the positions of the three remaining atoms
(depicted by two stars and a square), we would need a projection upto $\lmax  = 4$, because
the angle made by the `star', central atom and the `square' is $\frac{\pi}{2}$ radians.
However, if we used one channel to represent the `stars' and one to represent the `square',
we can get away by only using projections upto $\lmax = 2$, because the `stars' are
diametrically opposite each other.

\subsection{A Study on Learning Random Signals}
\label{sec:learning-random-signals}

To quantitatively show the effect of having multiple channels, we see how well the model is able to learn a random distribution on the sphere.
We randomly sample $N = 5$ target points with coordinates $\{\hat{\textbf{r}}_{i}\}_{i = 1}^N$ on the sphere, and then define the distribution:
$$
q(\hat{\textbf{r}}) = \sum_{i = 1}^N \exp(\delta_{\lmax}\left(\hat{\textbf{r}} - \hat{\textbf{r}}_{i} \right))
$$
with the same Dirac delta distribution approximation as described in \autoref{sec:delta-functions}. We use $\lmax = 5$ throughout this section.
Then, we randomly initialize coefficients $c$ to minimize the KL divergence to $q$:
$$
\min_c KL(q \ || \ p_c)
$$
where $p_c$ is the probability distribution defined by coefficients $c$, as before:
\begin{align*}
f(\theta, \phi) &= \log \sum_{\text{channel ch}} \exp \sum_{l = 0}^{\lmax} {c^{\text{ch}}_{l}}^T Y_l(\theta, \phi) \\
p(\theta, \phi) &= \frac{1}{Z}
 \exp{f(\theta, \phi)}
\end{align*}
This corresponds to a simpler setting where we have only one radius $r$.

\begin{figure}[!h]
    \centering
    \includegraphics[width=0.6\textwidth]{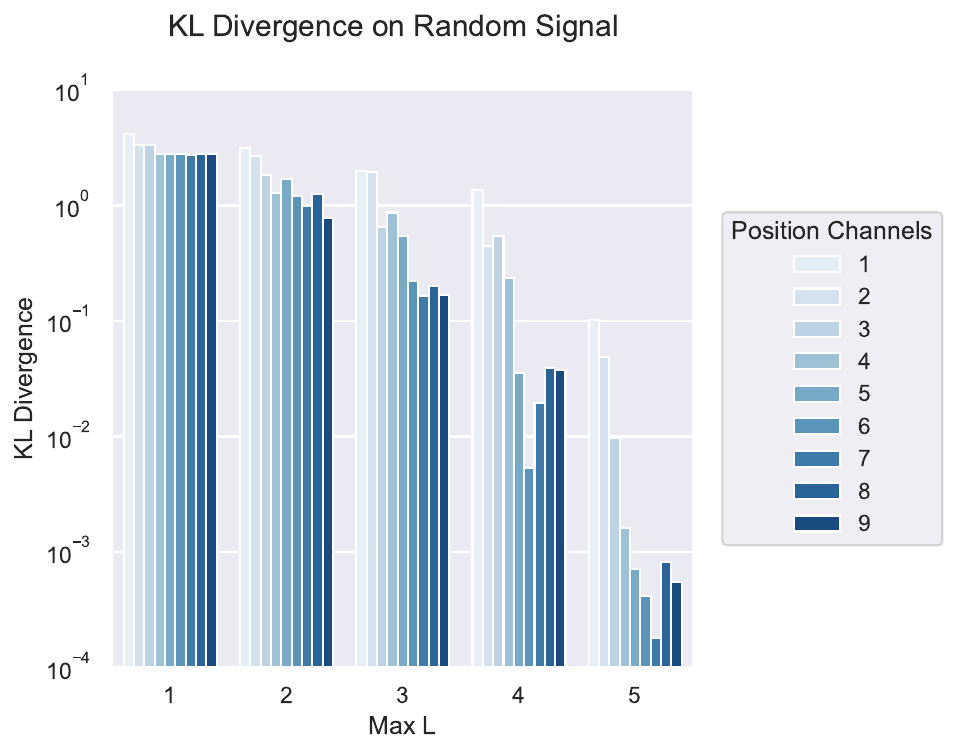}
    \caption{Final KL divergence $KL(q \ || \ p_c)$ for learned coefficients $c$ as a function of number of position channels $\text{ch}$  and $\lmax$.}
    \label{fig:kl_divergence_random_signal}
\end{figure}

We assess the KL divergence as a function of number of position channels $\text{ch}$ and $\lmax$ in \autoref{fig:kl_divergence_random_signal}. We see a consistent improvement across different $\lmax$ as the number of position channels are increased.

\begin{figure}[!h]
    \centering
    \includegraphics[width=\textwidth]{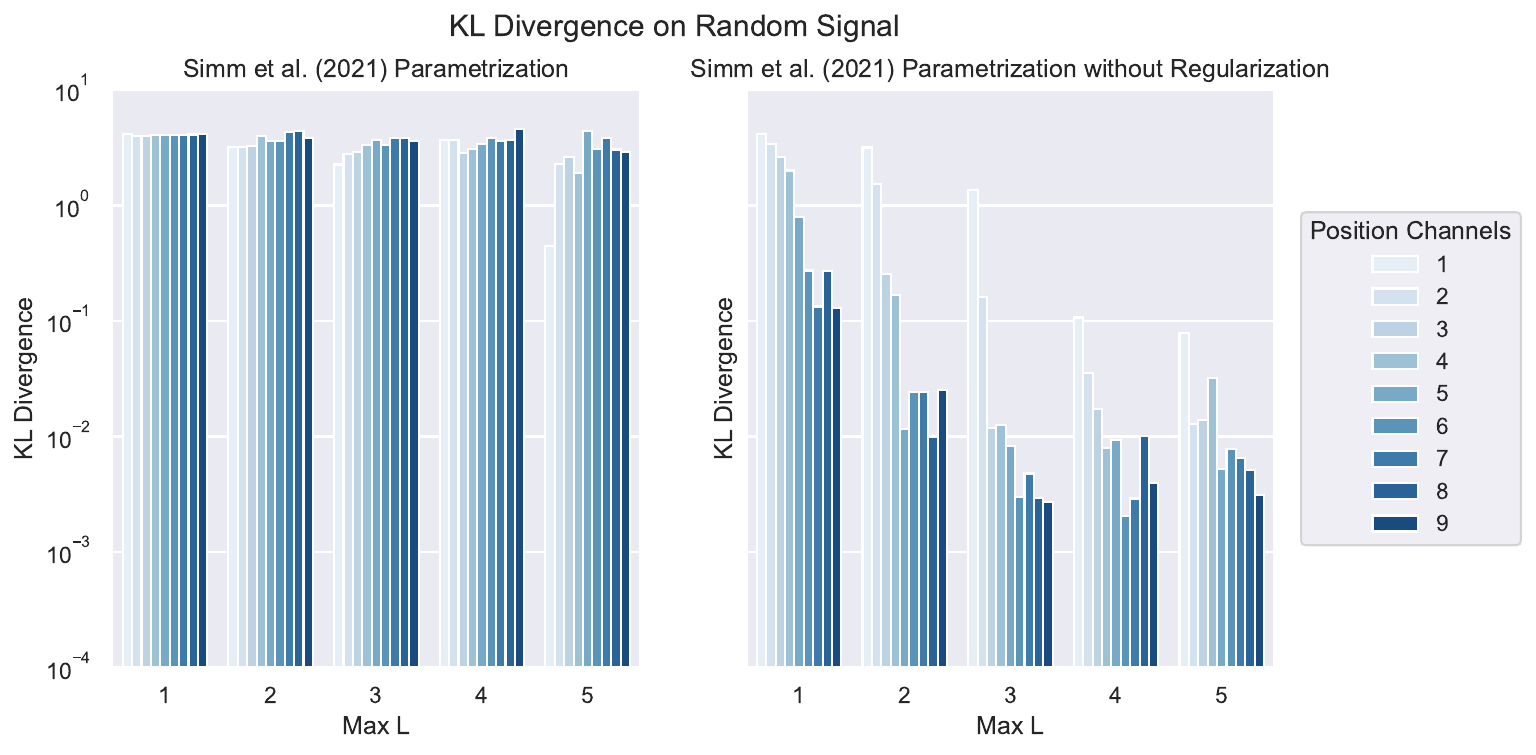}
    \caption{Final KL divergence $KL(q \ || \ p_c)$ for learned coefficients $c$ as a function of number of position channels $\text{ch}$  and $\lmax$, with the parametrization proposed by \citet{simm2021symmetryaware}. Removing the regularization term helps the model learn better.}
    \label{fig:kl_divergence_random_signal_simm}
\end{figure}

We also experimented with the parametrization from \citet{simm2021symmetryaware}, who define:
\begin{align*}
p(\theta, \phi) &= \frac{1}{Z}
 \exp{\left(-\frac{\beta}{k} |f(\theta, \phi)|^2\right)}
\end{align*}
where $k = \sum_{l = 0}^{\lmax} |c_l|^2$.
This extra factor of $k$ was proposed by \citet{simm2021symmetryaware}
to ``regularize the distribution so that it does not
approach a delta function".
In the left panel of \autoref{fig:kl_divergence_random_signal_simm}, we show that this
regularization hurts the model. Even adding multiple channels does not help,
because the regularization term `switches' off multiples 
channels. However, as shown in the right panel of 
\autoref{fig:kl_divergence_random_signal_simm}, removing this regularization 
significantly helps the model, with further improvement as the number of channels are 
increased.
For $\lmax = 5$, we see that our parametrization performs similarly to 
\citet{simm2021symmetryaware} without the regularization term.
Based on this experiment, we plan to 
experiment with non-linearities for the logits in future versions of Symphony.

\section{Ablation Studies}
\label{sec:ablation}

\subsection{$\lmax$ and Number of Position Channels}
To understand the practical effect of adding multiple position channels to Symphony, as well as the impact of increasing $\lmax$, we trained variants of Symphony varying
$\lmax$ for the focus embedder E3SchNet from $1$ to $2$, the number of position channels from $1$ to $4$, and $\lmax$ for the position embedder NequIP from $1$ to $5$.

Due to computational constraints, we trained these models for $1,000,000$ steps each, which is $8\times$ lesser than the original model reported in \autoref{sec:experiments}. Thus, the validity numbers are slightly lower overall. However, we believe we can still observe important trends from this experiment.

\begin{figure}[!h]
    \centering
    \includegraphics[width=\textwidth]{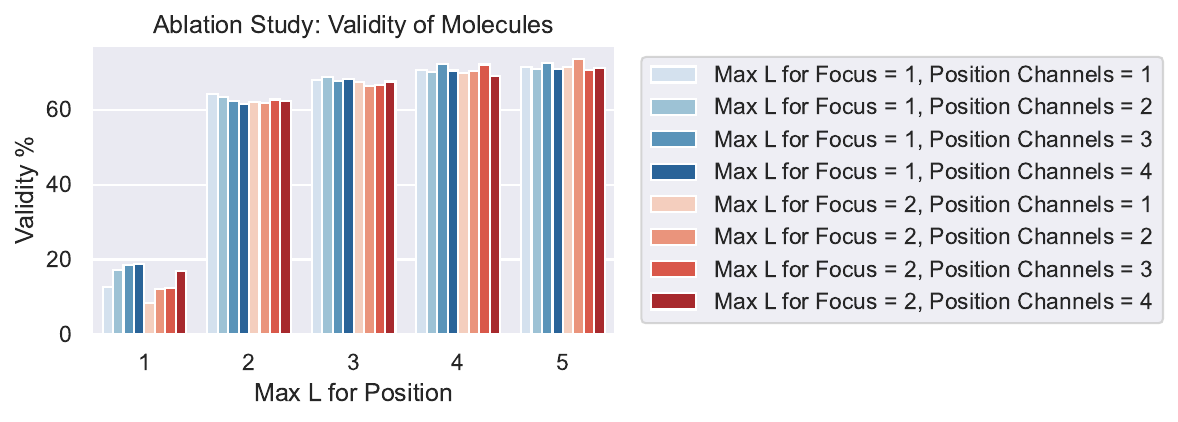}
    \caption{Validity as a function of $\lmax$ for the position and focus embedders. Models for which $\lmax = 1$ for the focus embedder are marked in blue.
    Models for which $\lmax = 2$ for the focus embedder are marked in red.
    The intensity of colours increases with the number of position channels.}
    \label{fig:validity_ablation}
\end{figure}

We report the validity as measured by \texttt{xyz2mol} for each of these models in \autoref{fig:validity_ablation}.

\begin{itemize}
    \item For the focus embedder E3SchNet, we do not see a significant increase in validity when going from $\lmax = 1$ to $\lmax = 2$.
    \item For the position embedder NequIP, we find a large jump when going from $\lmax = 1$ to $\lmax = 2$. Further increasing $\lmax$ seemed to help slightly. For computational reasons, we kept $\lmax = 5$.
    \item Increasing the number of position channels helps for $\lmax = 1$
    in particular.
\end{itemize}

\subsection{Resolution}
\label{sec:validity_vs_resolution}

\begin{figure}[h]
    \centering
    \includegraphics[width=\textwidth]{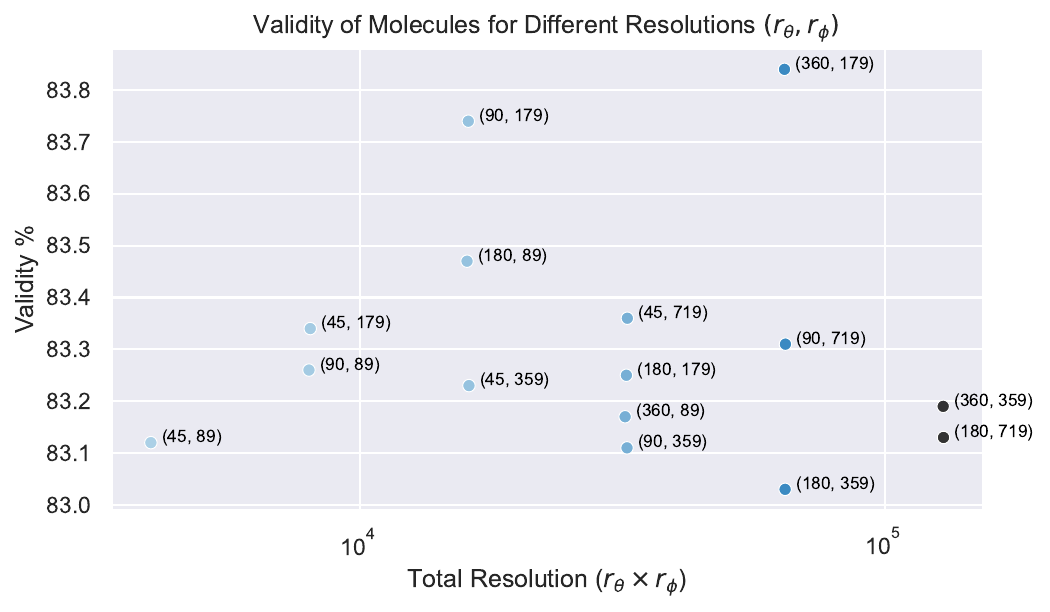}
    \caption{Validity as a function of sampling grid resolution $(r_\theta, r_\phi)$.}
    \label{fig:validity_vs_resolution}
\end{figure}

Here, we take the trained Symphony model, freeze all weights, and measure the validity of molecules across a range of grid resolutions.
The original grid resolution for model training was $(r_\theta, r_\phi) = (180, 359)$ as described above. From \autoref{fig:validity_vs_resolution}, we see that
the validity is within the expected variation even when using upto $10\times$ smaller grids. Further amplification of the resolution also
does not seem to affect the validity. We hypothesize that this is due to sampling with a lower temperature than ideal making the target distribution more diffuse; future work will seek to understand this effect better.

\begin{figure}[h]
    \centering
    \includegraphics[width=\textwidth]
    {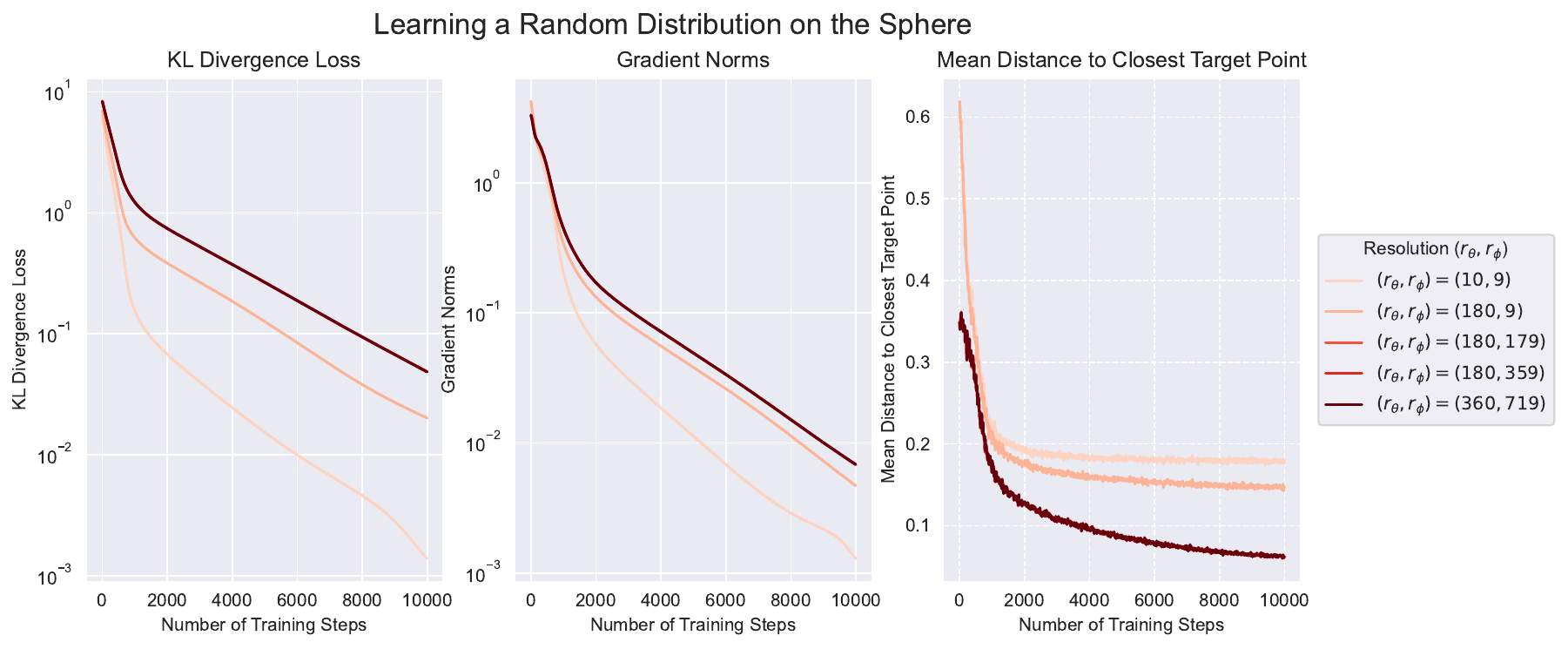}
    \caption{The effect of resolution when learning the random signal from \autoref{sec:learning-random-signals}. 
    Our original model was trained with a resolution of $(r_\theta, r_\phi) = (180, 359)$.}
    \label{fig:resolution_random_points}
\end{figure}

The previous experiment measured the effect of the grid resolution for sampling.
We also seeked to understand the effect of the grid resolution for training. For
this, we reuse the task of \autoref{sec:learning-random-signals}, and vary the
grid resolution.
All other hyperparameters were kept fixed, with $\lmax = 2$ and $2$ position
channels.
From \autoref{fig:resolution_random_points}, we see that the learning is not
affected even at low resolutions. In fact, from a KL divergence perspective, it is easier to learn at lower resolutions because localization is easier. However, lower resolutions come with decreased accuracy when sampling, as shown by the rightmost plot of \autoref{fig:resolution_random_points}.

\subsection{Temperature}
\label{sec:validity_vs_temperature}

\begin{figure}[p]
    \centering
    \includegraphics[width=0.8\textwidth]{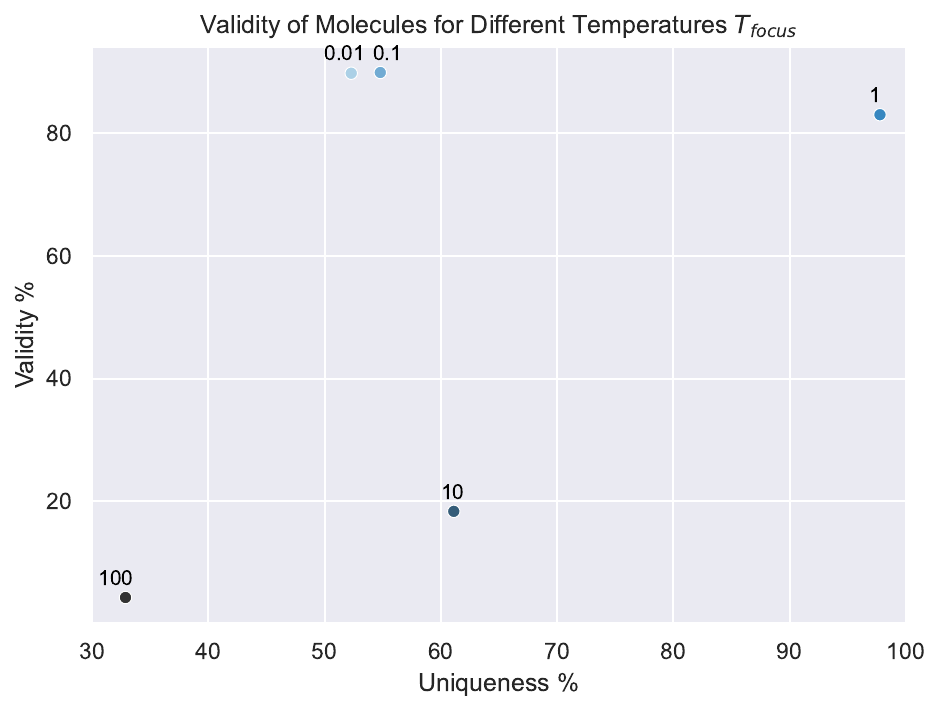}
    \includegraphics[width=0.8\textwidth]{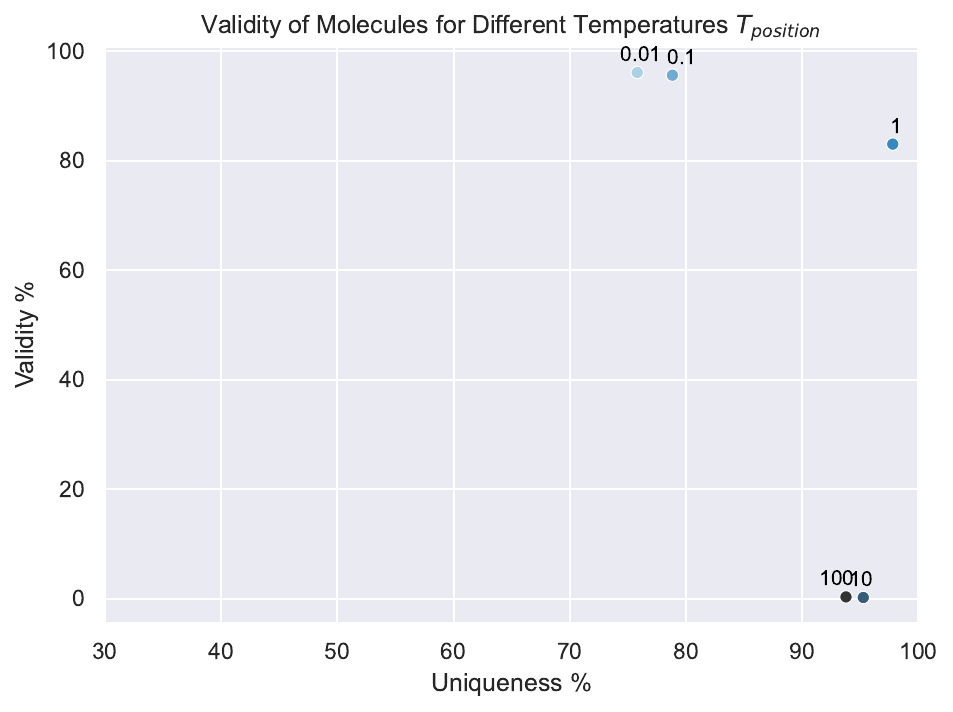}
    \caption{Validity as a function of temperature applied to the focus (above) and position (below) distribution logits.}
    \label{fig:validity_vs_temperature}
\end{figure}

Again, we take the trained Symphony model, freeze all weights,
and measure the validity
of molecules across a range of temperatures $T$.
This means scaling all the logits by 
a factor of $\frac{1}{T}$. Higher temperatures make the model more diffuse, while 
lower temperatures make the model more peaked.
We see that while the validity improves
significantly at lower temperatures, the uniqueness tends to suffer.
As seen in \autoref{fig:validity_vs_temperature}, this 
experiment suggests a more careful sampling of the temperature to better understand a
Pareto frontier between validity and uniqueness.

\section{Representing Dirac Delta Distributions}
\label{sec:delta-functions}

Suppose we have the function $f(\hat{\textbf{r}}) = \delta(\hat{\textbf{r}} - \hat{\textbf{r}}_0)$ defined on the sphere $S^2$, and we wish to compute its spherical harmonic coefficients $c_{l,m}$:
\begin{align*}
   f(\theta, \phi) = \sum_{l = 0}^{\lmax} c_l^T Y_l(\theta, \phi) = \sum_{l = 0}^{\lmax} \sum_{m = -l}^{l} c_{l,m} Y_{l,m}(\theta, \phi) 
\end{align*}
By orthonormality of the spherical harmonics, and the annihilation property of the Dirac delta:
\begin{align*}
   c_{l,m} &= \int f(\theta, \phi)  Y_{l,m}(\theta, \phi) \sin \theta d\theta d\phi \\
   &= \int \delta(\hat{\textbf{r}} - \hat{\textbf{r}}_0)  Y_{l,m}(\theta, \phi) \sin \theta d\theta d\phi \\
   &= Y_{l,m}(\hat{\textbf{r}}_0)
\end{align*}
Thus, we can easily compute the spherical harmonic coefficients for the Dirac delta distribution upto any required $\lmax$. This is implemented in the \texttt{e3nn-jax} package. Due to the frequency cutoff, the Dirac delta distribution thus obtained is a smooth approximation of a true Dirac delta. 

\section{The Effect of Distance Cutoffs on Diffusion Models}
\label{sec:edm-radial-cutoff}

As described in \autoref{sec:related}, EDM \citep{edm} and other diffusion models use fully-connected graphs to denoise the 3D molecular graph at each timestep.
Here, we investigate the performance of EDM as we apply a distance-based cutoff $c$ to the edges:
$$
    (i, j) \in E \iff || \r_i - \r_j || \leq c.
$$

In \autoref{tab:num_edges_radial_cutoff}, we vary $c$ on the QM9 dataset to measure how many of the edges in the fully-connected graphs are kept after the radial cutoff.

\begin{table}[h]
\centering
    \begin{tabular}{cc}
    \toprule
    Radial Distance Cutoff & Percentage of Edges Preserved  \\
    \midrule
    $2.0$ A & $\approx 26\%$ \\
    $3.0$ A & $\approx 61\%$ \\
    $5.0$ A & $\approx 93\%$ \\
    $10.0$ A& $100\%$ \\
    \end{tabular}
    \caption{Percentage of edges preserved by the radial distance cutoff for molecular graphs from QM9.}
    \label{tab:num_edges_radial_cutoff}
\end{table}

\begin{figure}[h]
    \centering
    \includegraphics[width=0.8\textwidth]{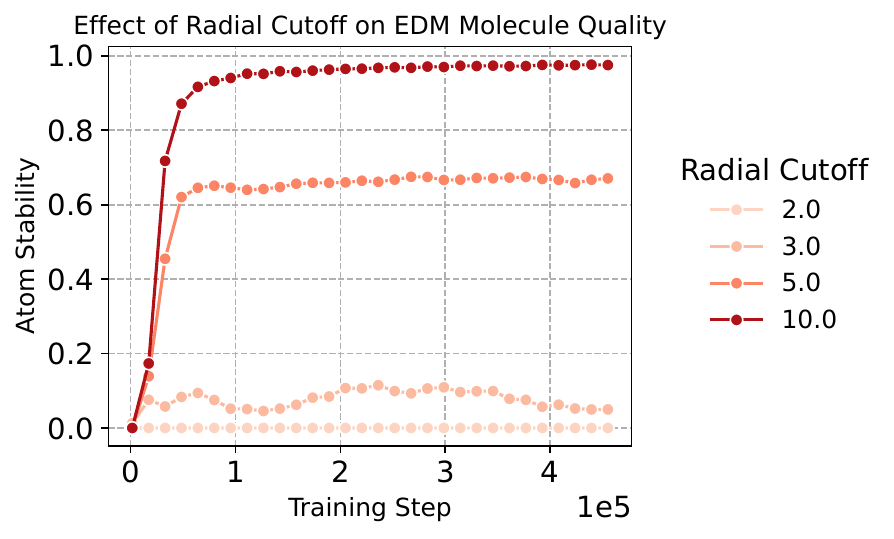}
    \caption{Atom stability on QM9 as measured by \citet{edm} as a function of radial cutoff $c$, showing that the model is quite sensitive.}
    \label{fig:edm-radial-cutoffs}
\end{figure}

In \autoref{fig:edm-radial-cutoffs}, we measured the `atom stability' as computed by 
\citet{edm} on the EDM-generated molecules as $c$ is varied, when trained on QM9. We notice a sharp drop-off in this metric as $c$ (and hence, the number of edges) reduces, despite the use of many message-passing layers in the underlying graph neural network.
Even for $c = 5$ A, where $93\%$ of the edges in the graph still remain, the quality of the EDM-generated molecules drops significantly. On observing the trajectories of the individual atoms, we noticed that the system tends to explode for lower values of $c$. Note that Symphony uses a radial cutoff of $5$ A as well.

\section{Generated Molecules From Symphony}
\begin{figure}[htbp]
    \centering
    \includegraphics[width=0.24\textwidth]{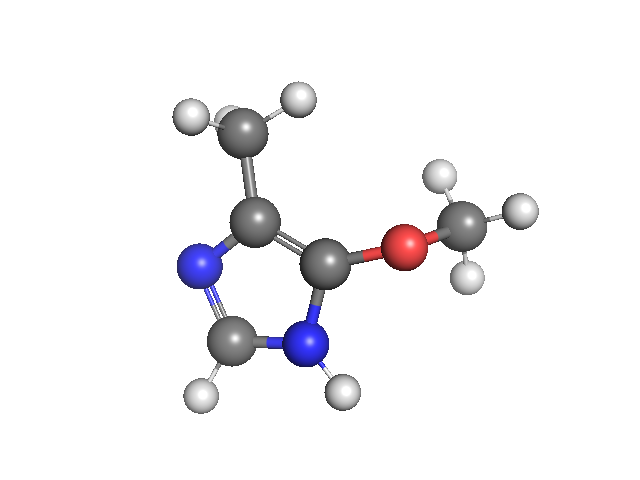}
    \includegraphics[width=0.24\textwidth]{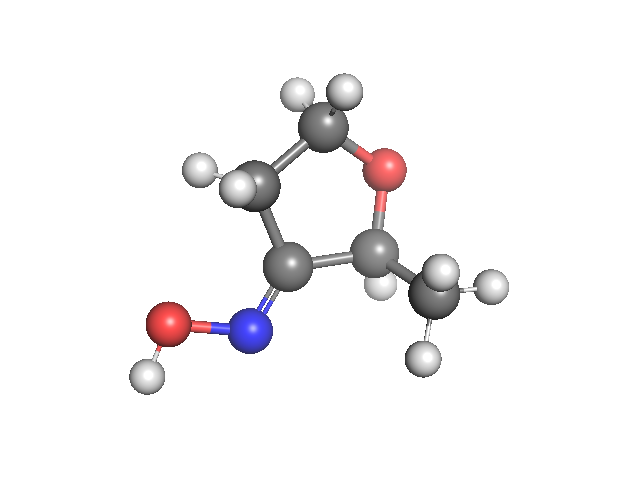}
    \includegraphics[width=0.24\textwidth]{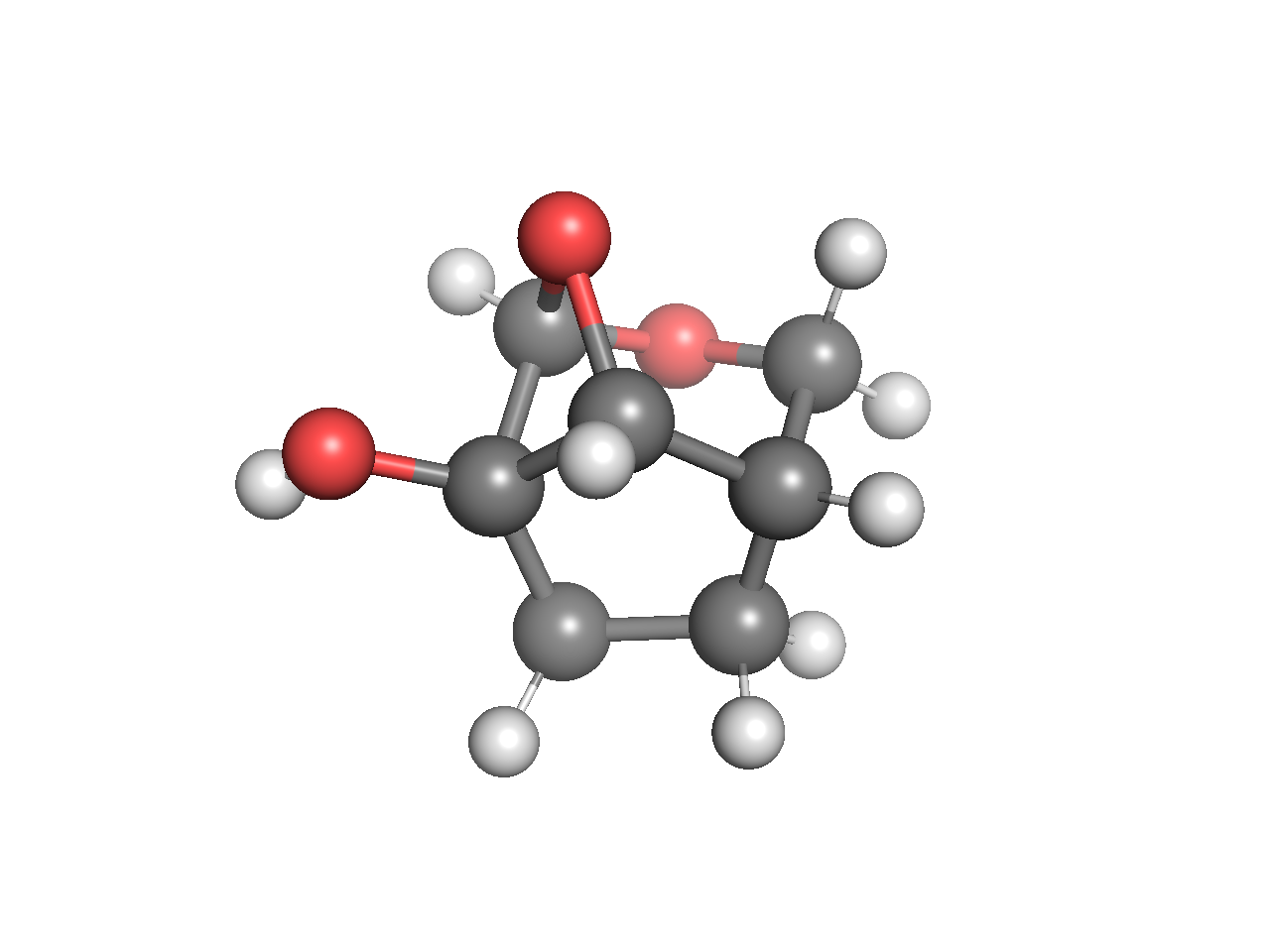}
    \includegraphics[width=0.24\textwidth]{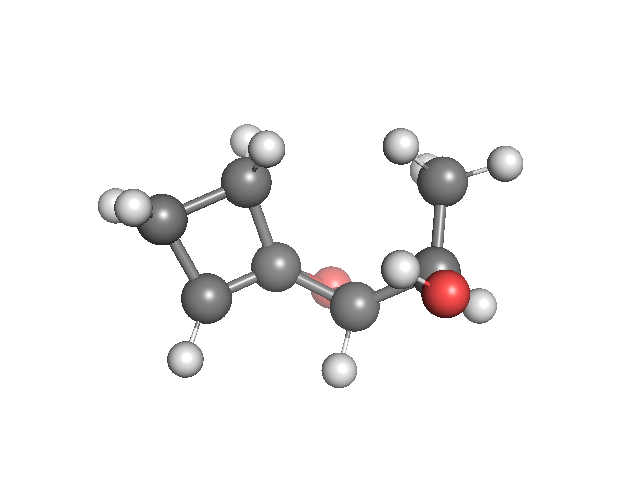}
    \includegraphics[width=0.24\textwidth]{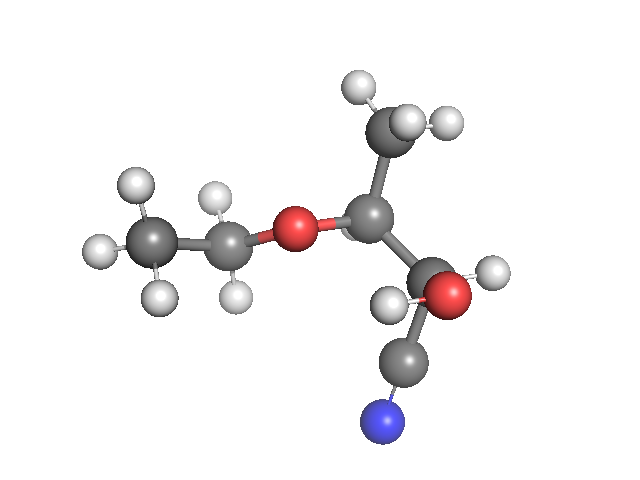}
    \includegraphics[width=0.24\textwidth]{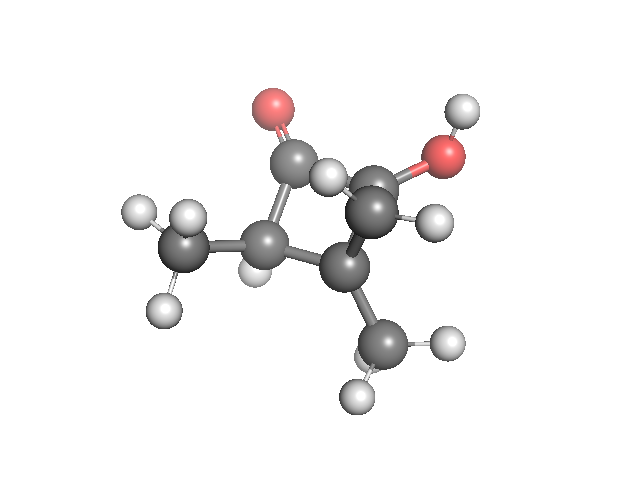}
    \includegraphics[width=0.24\textwidth]{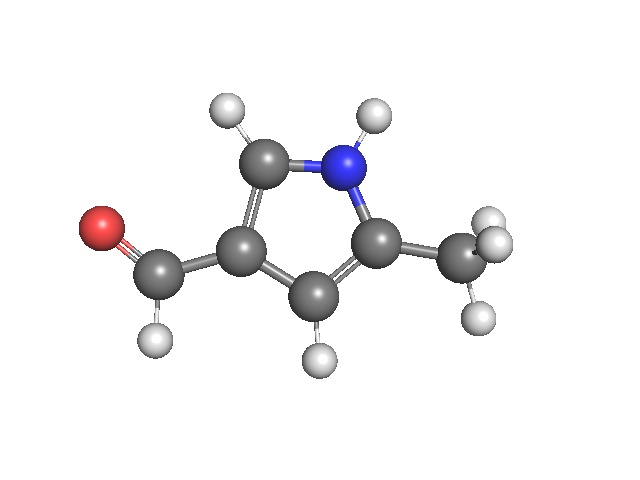}
    \includegraphics[width=0.24\textwidth]{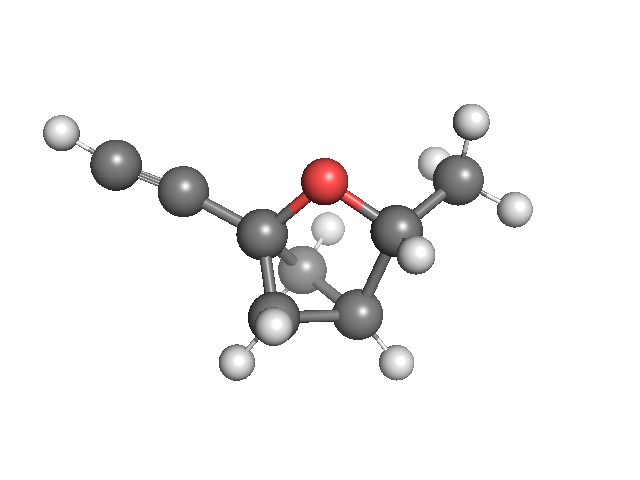}
    \includegraphics[width=0.24\textwidth]{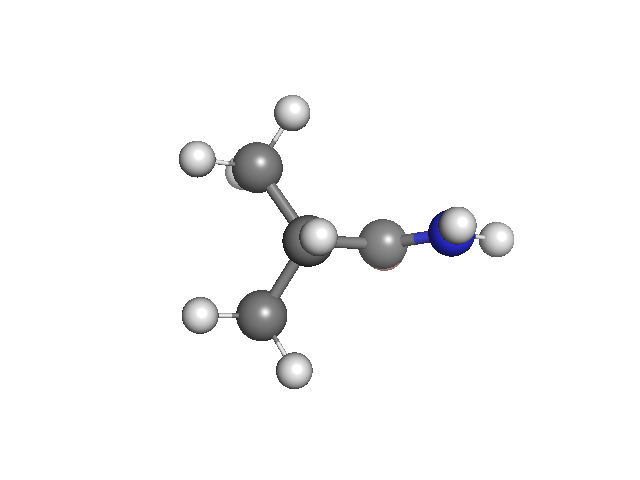}
    \includegraphics[width=0.24\textwidth]{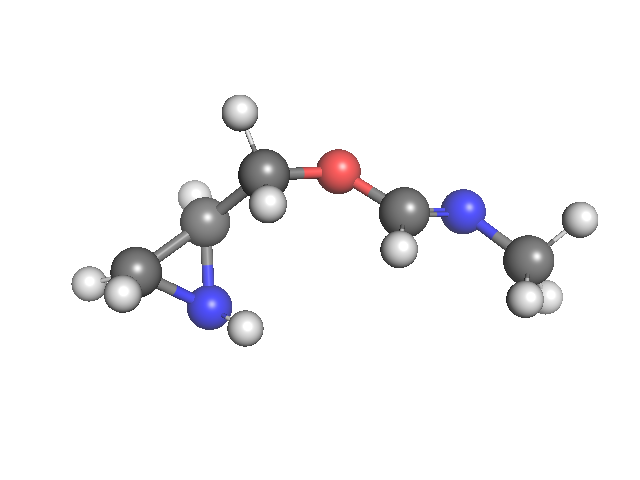}
    \includegraphics[width=0.24\textwidth]{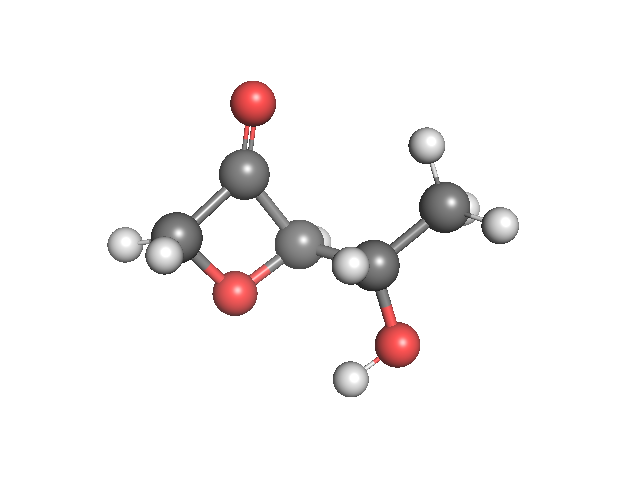}
    \includegraphics[width=0.24\textwidth]{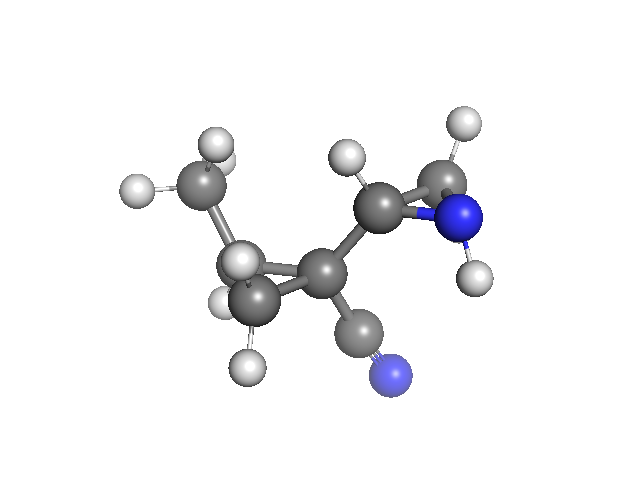}
    \includegraphics[width=0.24\textwidth]{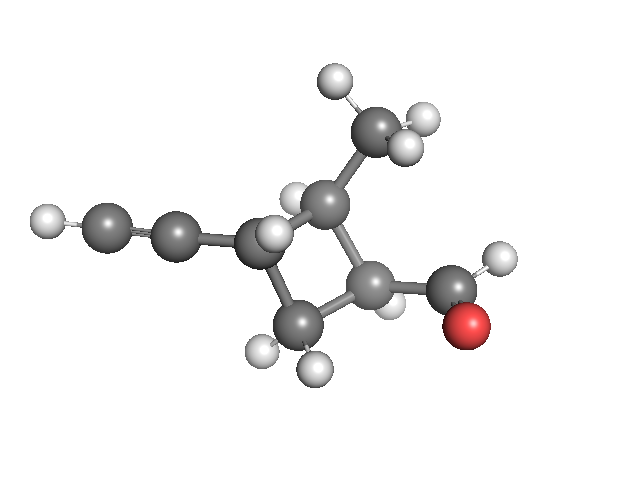}
    \includegraphics[width=0.24\textwidth]{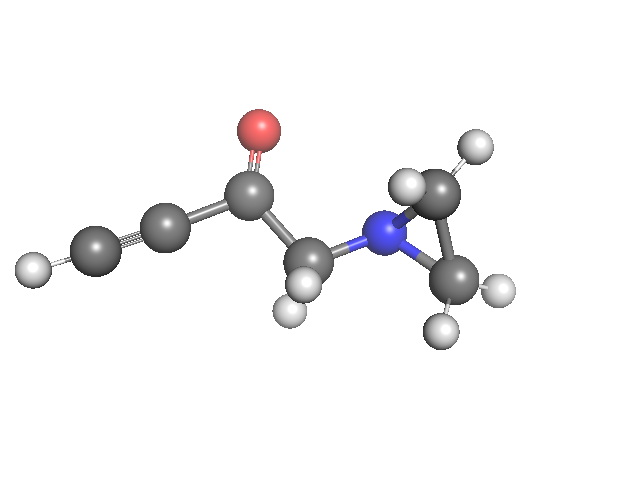}
    \includegraphics[width=0.24\textwidth]{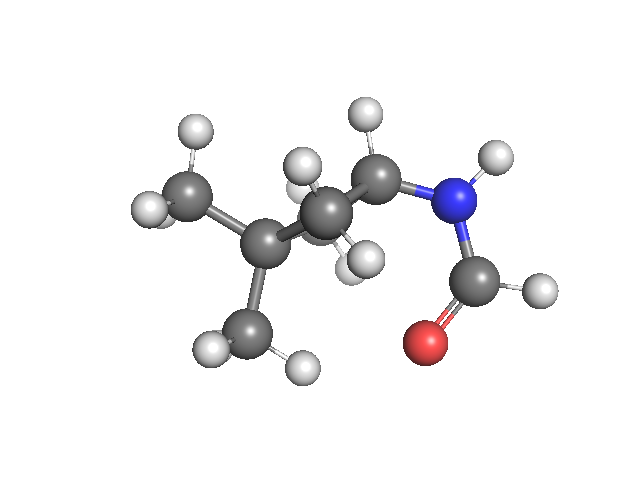}
    \includegraphics[width=0.24\textwidth]{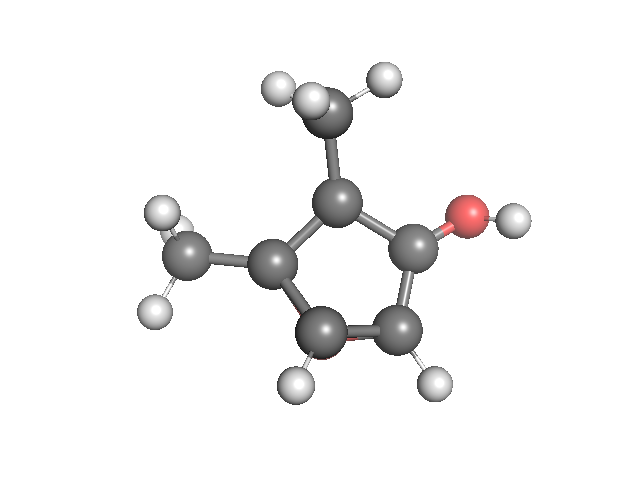}
    \includegraphics[width=0.24\textwidth]{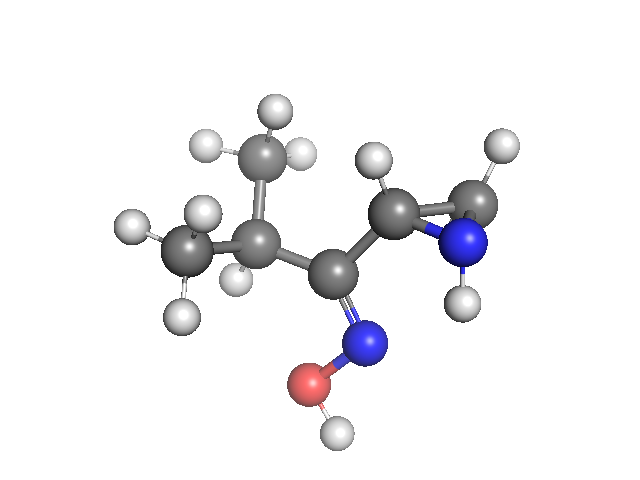}
    \includegraphics[width=0.24\textwidth]{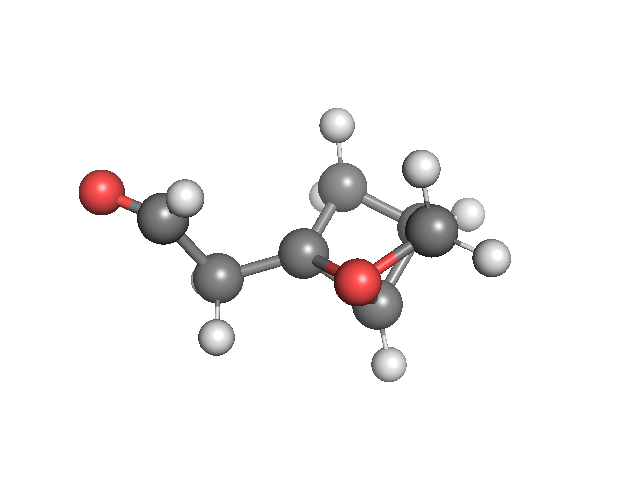}
    \includegraphics[width=0.24\textwidth]{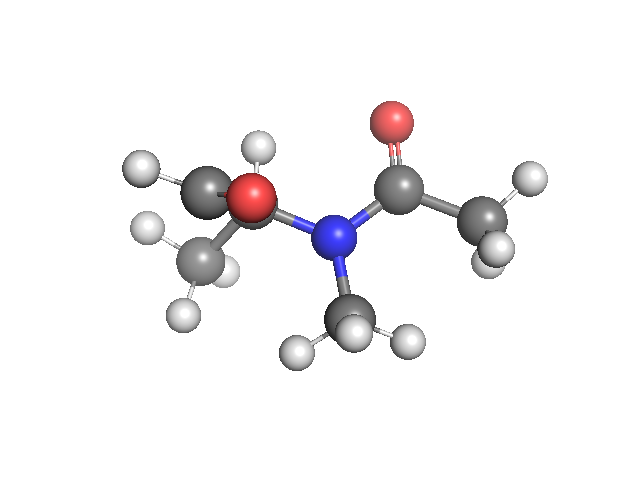}
    \includegraphics[width=0.24\textwidth]{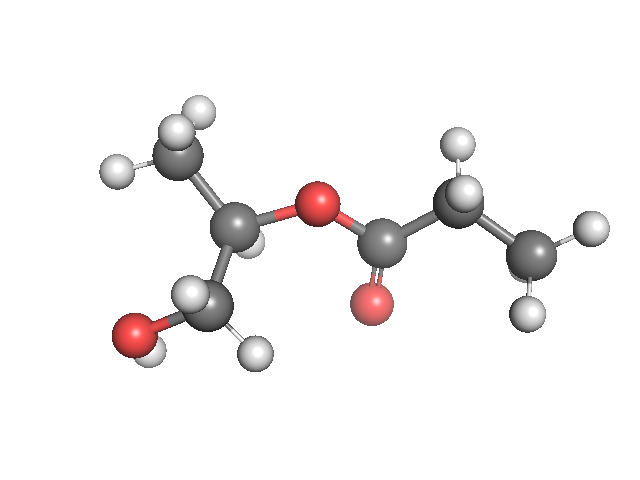}
    \includegraphics[width=0.24\textwidth]{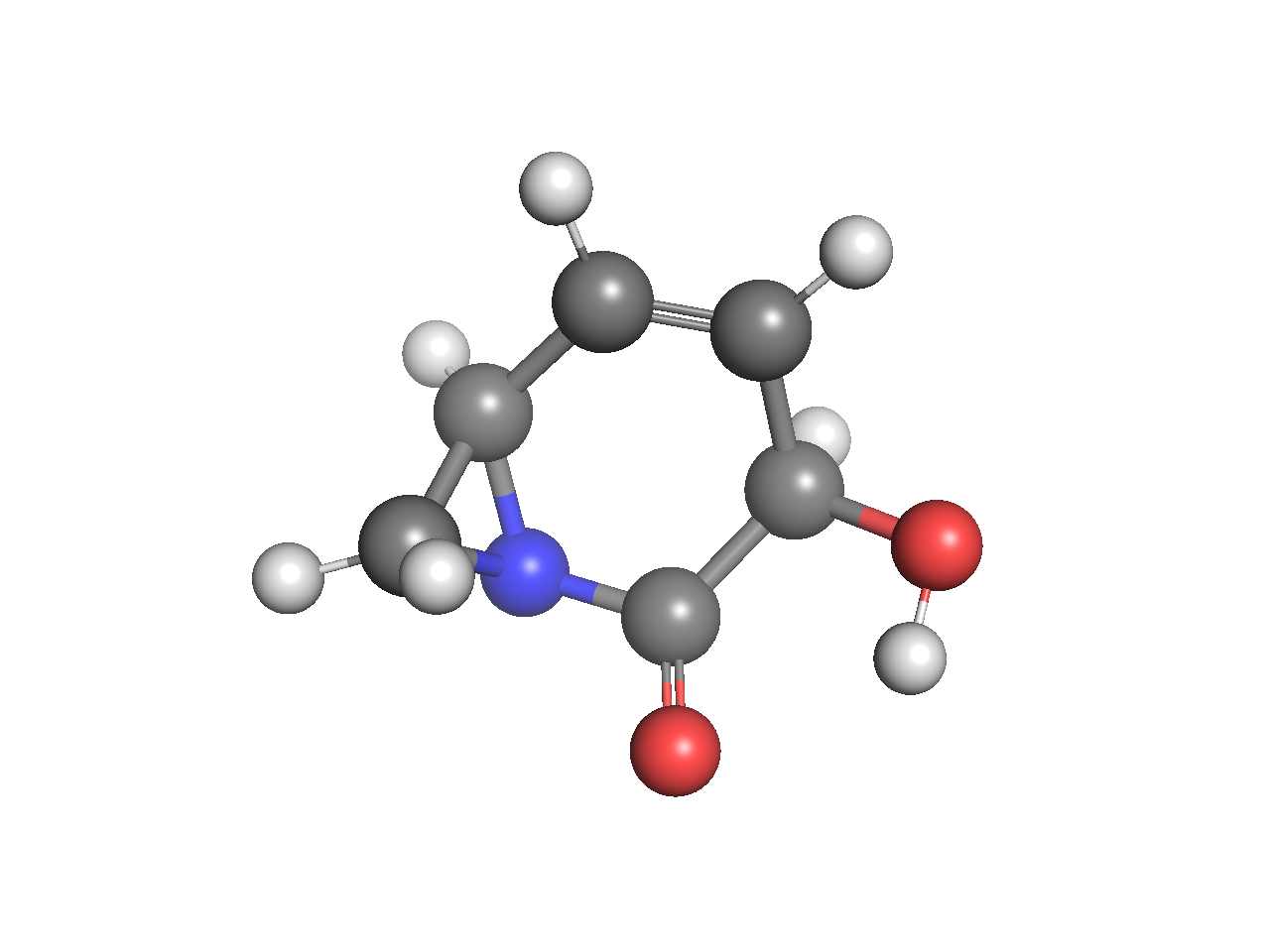}
    \includegraphics[width=0.24\textwidth]{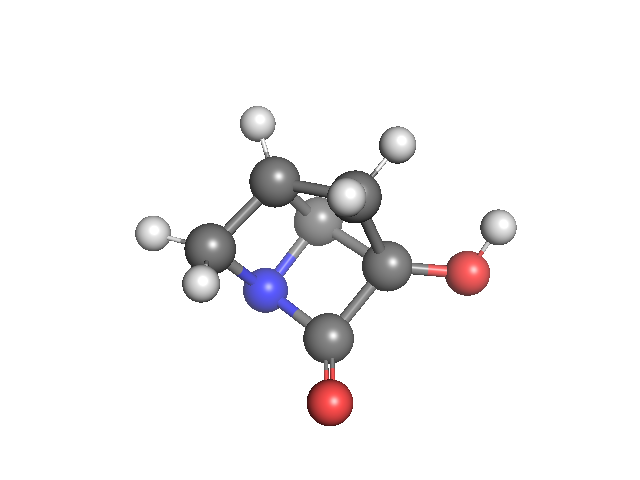}
    \includegraphics[width=0.24\textwidth]{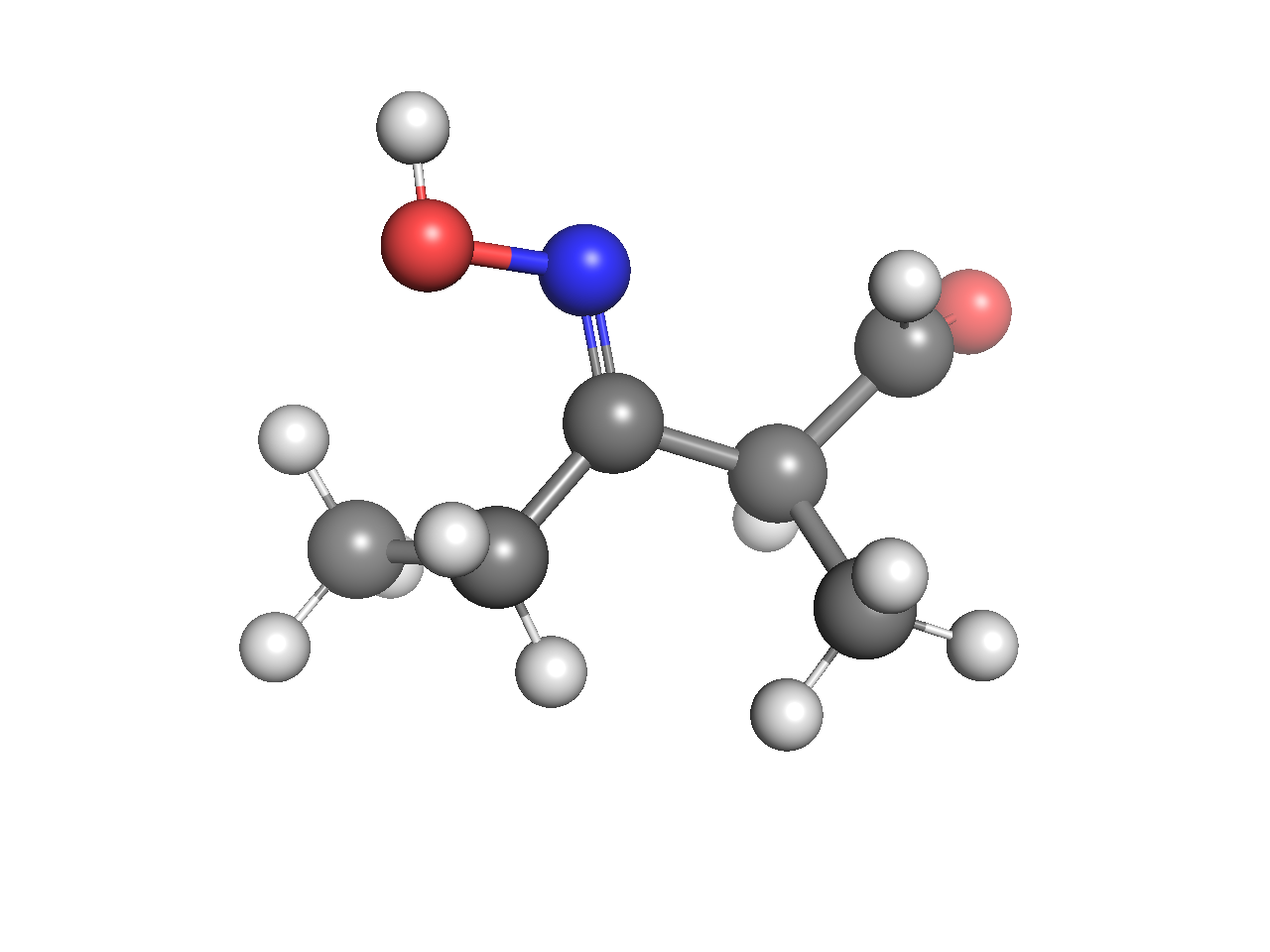}
    \includegraphics[width=0.24\textwidth]{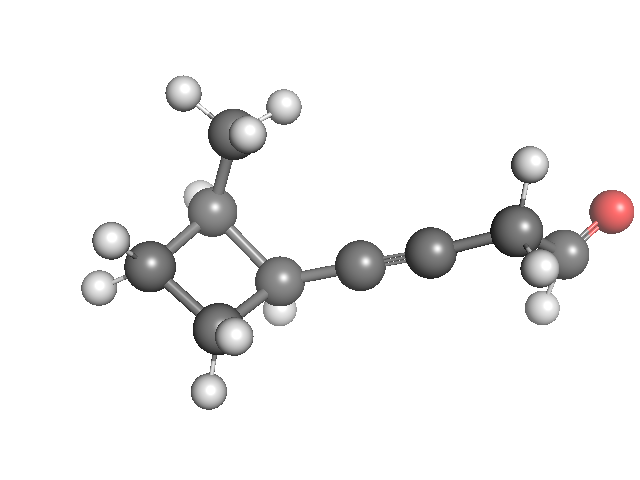}
    \caption{Molecules generated by Symphony and visualized with PyMOL \citep{PyMOL}.}
    \label{fig:generated_mols}
\end{figure}

\autoref{fig:generated_mols} exhibits random non-cherry-picked samples from Symphony.

\end{document}